\documentclass[twocolumn, switch]{article} 

\usepackage{preprint}

\usepackage{amsmath, amsthm, amssymb, amsfonts}

\usepackage[numbers,square]{natbib}
\usepackage[utf8]{inputenc}	
\usepackage[T1]{fontenc}	
\usepackage{xcolor}		
\usepackage[colorlinks = true,
            linkcolor = purple,
            urlcolor  = blue,
            citecolor = cyan,
            anchorcolor = black]{hyperref}	
\usepackage{booktabs} 		
\usepackage{nicefrac}		
\usepackage{microtype}		
\usepackage{lineno}		
\usepackage{float}			

\usepackage{lipsum}		

\usepackage{newfloat}
\DeclareFloatingEnvironment[name={Supplementary Figure}]{suppfigure}
\usepackage{sidecap}
\sidecaptionvpos{figure}{c}

\usepackage{titlesec}
\titlespacing\section{0pt}{12pt plus 3pt minus 3pt}{1pt plus 1pt minus 1pt}
\titlespacing\subsection{0pt}{10pt plus 3pt minus 3pt}{1pt plus 1pt minus 1pt}
\titlespacing\subsubsection{0pt}{8pt plus 3pt minus 3pt}{1pt plus 1pt minus 1pt}

\usepackage{tikz,xcolor,hyperref}

\definecolor{lime}{HTML}{A6CE39}
\DeclareRobustCommand{\orcidicon}{
	\begin{tikzpicture}
	\draw[lime, fill=lime] (0,0) 
	circle [radius=0.16] 
	node[white] {{\fontfamily{qag}\selectfont \tiny ID}};
	\draw[white, fill=white] (-0.0625,0.095) 
	circle [radius=0.007];
	\end{tikzpicture}
	\hspace{-2mm}
}
\foreach \x in {A, ..., Z}{\expandafter\xdef\csname orcid\x\endcsname{\noexpand\href{https://orcid.org/\csname orcidauthor\x\endcsname}
			{\noexpand\orcidicon}}
}

\title{A Review on Deep Learning in UAV Remote Sensing}

\usepackage{authblk}

\author[1\thanks{\tt{lucasosco@unoeste.br}}]{Lucas Prado Osco\orcidA{}}
\author[2]{José Marcato Junior\orcidB{}}
\author[3]{Ana Paula Marques Ramos\orcidC{}}
\author[4]{Lúcio André de Castro Jorge\orcidD{}}
\author[5]{Sarah Narges Fatholahi\orcidE{}}
\author[6]{Jonathan de Andrade Silva\orcidF{}}
\author[6]{Edson Takashi Matsubara\orcidG{}}
\author[7]{Hemerson Pistori\orcidH{}}
\author[6]{Wesley Nunes Gonçalves\orcidI{}}
\author[5]{Jonathan Li\orcidJ{}}

\affil[1]{\scriptsize Faculty of Engineering and Architecture and Urbanism, University of Western São Paulo (UNOESTE), Rod. Raposo Tavares, km 572, Limoeiro, Presidente Prudente 19067-175, SP, Brazil; lucasosco@unoeste.br; pradoosco@gmail.com}
\affil[2]{Faculty of Engineering, Architecture and Urbanism and Geography, Federal University of Mato Grosso do Sul (UFMS), Av. Costa e Silva-Pioneiros, Cidade Universitária, Campo Grande 79070-900, MS, Brazil; jose.marcarto@ufms.br}
\affil[3]{Departament of Cartography, São Paulo State University (UNESP), Centro Educacional, R. Roberto Simonsen, 305, Presidente Prudente, 19060-900, SP, Brazil; marques.ramos@unesp.br}
\affil[4]{National Research Center of Development of Agricultural Instrumentation, Brazilian Agricultural Research Agency, R. XV de Novembro, 1452, São Carlos, 13560-970, SP, Brazil; lucio.jorge@embrapa.br}
\affil[5]{Department of Geography and Environmental Management, University of Waterloo, Waterloo, ON N2L 3G1, Canada; nfatholahi@uwaterloo.ca, junli@uwaterloo.ca}
\affil[6]{Faculty of Computing, Federal University of Mato Grosso do Sul (UFMS), Av. Costa e Silva-Pioneiros, Cidade Universitária, Campo Grande 79070-900, MS, Brazil; jonathan.andrade@ufms.br, edsontm@facom.ufms.br, wesley.goncalves@ufms.br}
\affil[7]{Inovisão, Catholic University of Dom Bosco, Av. Tamandaré, 6000, Campo Grande, 79117-900, MS, Brazil; pistori@ucdb.br}

\begin{document}

\twocolumn[ 
  \begin{@twocolumnfalse} 
  
\maketitle

\begin{abstract}
\small Deep Neural Networks (DNNs) learn hierarchical representations from data, bringing significant advances in image processing, and time-series analysis, as well as in natural language, audio, video, and many others. In the field of remote sensing, research and literature reviews specifically involving DNN applications have been conducted to summarize the amount of information produced. Recently, applications based on Unmanned Aerial Vehicles (UAVs) have stood out in aerial sensing research, as they allow for fast, less costly data collection at high spatial resolution. However, a literature review that combines the themes of "Deep Learning" (DL) and "remote sensing with UAVs" has not yet been conducted. The motivation for our work was to present a review of the fundamentals of DL applied to images collected by sensors onboard these aircraft. We especially present a description of the classification and segmentation techniques used in recent applications with data acquired by UAVs. For this, a total of 232 articles published in international scientific journal databases were examined. We gathered all this material and evaluated its characteristics in relation, for example, to the application, sensor, and type of network used. We relate how DL presents promising results and has the potential for processing tasks associated with aerial image data collected by UAVs. Finally, we project future perspectives, commenting on the prominent paths of DL to be explored in aerial remote sensing. Our review consists of a simplistic and objective approach to present, comment and summarize the state of the art in applications of sub-meter spatial resolution images with DNNs in various subfields of remote sensing, grouping them in the environmental, urban, and agricultural contexts. 
\end{abstract}
\vspace{0.35cm}

  \end{@twocolumnfalse} 
] 



\section{Introduction}

For investigations using remote sensing image data, multiple processing tasks depend on computer vision algorithms. In the past decade, applications conducted with statistical and Machine Learning (ML)  algorithms were mainly used in classification/regression tasks. The increase of remote sensing systems allowed a wide collection of data from any target on the Earth’s surface. Aerial imaging has become a common approach to acquiring data with the advent of Unnamed Aerial Vehicles (UAV). These are also known as Remotely Piloted Aircrafts (RPA), or, as a commonly adopted term, drones (multi-rotor, fixed wings, hybrid, etc). These devices have grown in market availability for their relatively low cost and high operational capability to capture images quickly and in an easy manner. The high-spatial-resolution of UAV-based imagery and its capacity for multiple visits allowed the creation of large and detailed amounts of datasets to be dealt with.

The surface mapping with UAV platforms presents some advantages compared to orbital and other aerial sensing methods of acquisition. Less atmospheric interference, the possibility to fly within lower altitudes, and mainly, the low operational cost have made this acquisition system popular in both commercial and scientific explorations. However, the visual inspection of multiple objects can still be a time-consuming, biased, and inaccurate operation. Currently, the real challenge in remote sensing approaches is to obtain automatic, rapid, and accurate information from this type of data. In recent years, the advent of Deep Learning (DL) techniques has offered robust and intelligent methods to improve the mapping of the Earth’s surface.

DL is an Artificial Neural Network (ANN) method with multiple hidden layers and deeper combinations, which is responsible for optimizing and returning better learning patterns than a common ANN. There is an impressive amount of revision material in the scientific journals explaining DL-based techniques, its historical evolution, general usage, as well as detailing networks and functions. Highly detailed publications, such as Lecun \citep{Lecun2015} and Goodfellow \citep{Goodfellow2016} are both considered important material in this area. As computer processing and labeled examples (i.e. samples) became more available in recent years, the performance of Deep Neural Networks (DNNs) increased in the image-processing applications. DNN has been successfully applied in data-driven methods. However, much needs to be covered to truly understand its potential, as well as its limitations. In this regard, several surveys on the application of DL in remote sensing were developed in both general and specific contexts to better explain its importance.

The context in which remote sensing literature surveys are presented is variated. Zhang et al. \citep{Zhang2016} organized a revision material which explains how DL methods were being applied, at the time, to image classification tasks. Later, Cheng et al. \citep{Cheng2016} investigated object detection in optical images, but focused more on the traditional ANN and ML. A complete and systematic review was presented by Ball et al. \citep{Ball2017} in a survey describing DL theories, tools, and its challenges in dealing with remote sensing data. Cheng et al. \citep{Cheng2017} produced a revision on image classification with examples produced at their experiments. Also, focusing on classification, Zhu et al. \citep{Zhu2017} summarized most of the current information to understand the DL methods used for this task. Additionally, a survey performed by Li et al. \citep{Li2018a} helped to understand some DL applications regarding the overall performance of DNNs in publicly available datasets for image classification task. Yao et al. \citep{Yao2018} stated in their survey that DL will become the dominant method of image classification in remote sensing community.

Although DL does provide promising results, many observations and examinations are still required. Interestingly enough, multiple remote sensing applications using hyperspectral imagery (HSI) data were in the process, which gained attention. In Petersson et al. \citep{Petersson2017}, probably one of the first surveys on hyperspectral data was performed. In \citep{Signoroni2019}, is presented a multidisciplinary review about how DL models have been widely used in the field of HSI dataset processing. These authors highlighted that, among the distinct areas of applications, remote sensing approaches are one of the most emerging. Regarding the use of DL models to process highly detailed remotely sensed HSI data, Signoroni et al. \citep{Signoroni2019} summarized usage into classification tasks, object detection, semantic segmentation, and data enhancement, such as denoising, spatial super-resolution, and fusion. Adão et al. \citep{Adao2020} present a recent review on hyperspectral imaging acquired by UAV-based sensors for agriculture and forestry applications, and show that there are manifold DL approaches to deal with HSI dataset complexity.

A more recent survey is presented by Jia et al. \citep{JIA2021} regarding DL for hyperspectral image classification considering few labeled samples. They commentate how there is a notable gap between deep learning models and HSI datasets because DL models usually need sufficient labeled samples, but it is generally difficult to acquire many samples in HSI dataset due to the difficulty and time-consuming nature of manual labeling. However, the issues of small-sample sets may be well defined by the fusion of deep learning methods and related techniques, such as transfer learning and a lightweight model. Deep learning is also a new approach for the domain of infrared thermal imagery processing to attend different domains, especially in satellite-provided data. Some of these applications are the usage of convolutional layers to detect potholes on roads with terrestrial imagery \citep{APARNA2019}, detection of land surface temperatures from combined multispectral and microwave observations from orbital platforms \citep{Wang2020lst}, or determining sea surface temperature patterns to identify ocean temperatures extremes \citep{XavierProchaska2021} from orbital imagery.

Yet in the literature revision theme, a comparative review by Audebert et al. \citep{Audebert2019} was conducted by examining various families of networks' architectures while providing a toolbox  to perform such methods to be publicly available. In this regard, another paper written by Paoletti et al. \citep{Paoletti2019} organized the source code of DNNs to be easily reproduced. Similar to \citep{Cheng2017}, Li et al. \citep{Li2019} conducted a literature revision while presenting an experimental analysis with DNNs' methods. As of recently, literature revision focused on more specific approaches within this theme. Some of which included DL methods for enhancement of remote sensing observations, as super-resolution, denoising, restoration, pan-sharpening, and image fusion techniques, as demonstrated by Tsagkatakis et al. \citep{Tsagkatakis2019} and Signoroni et al. \citep{Signoroni2019}. Also, a meta-analysis by Ma et al. \citep{Ma2019} was performed concerning the usage of DL algorithms in seven subfields of remote sensing: image fusion and image registration, scene classification, object detection, land use and land cover classification, semantic segmentation, and object-based image analysis (OBIA). 

Although, from these recent reviews, various remote sensing applications using DL can be verified, it should be noted that the authors did not focus on specific surveying in the context of DL algorithms applied to UAV-image sets, which is something that, at the time of writing, has gained the attention of remote sensing investigations. We verified in the literature that, in general, similar DL methods are used for imagery acquired at different levels, resolutions and domains, such as the ones from orbital, aerial, terrestrial and proximal sensing platforms. However, as of recently, some of the proposed deep neural networks are maintaining high resolution images into deeper layers \citep{Kannojia2018}. This type of deep networks may benefit from UAV-based data, taking advantage of its resolutions. Indeed, there are orbital images with high spatial resolutions, but these are not as commonly available to the general public as UAV-based images. Because of that, these kinds of architectures associated with UAV-based data may be a surging trend in remote sensing applications.

Another interesting take on DL-based methods was related to image segmentation in a survey by Hossain et al. \citep{Hossain2019}, which its theme was expanded by Yuan et al. \citep{Yuan2021} and included state-of-the-art algorithms. A summarized analysis by Zheng et al.  \citep{Zheng2020a} focused on remote sensing images with object detection approaches, indicating some of the challenges related to the detection with few labeled samples, multi-scale issues, network structure problems, and cross-domain detection difficulties. In more of a ``niche'' type of research, environmental applications and land surface change detection were investigated in literature revision papers by Yuan et al. \citep{Yuan2020} and Khelifi et al. \citep{Khelifi2020}, respectively.

\begin{figure}[ht!]
\centering
\includegraphics[width=\columnwidth]{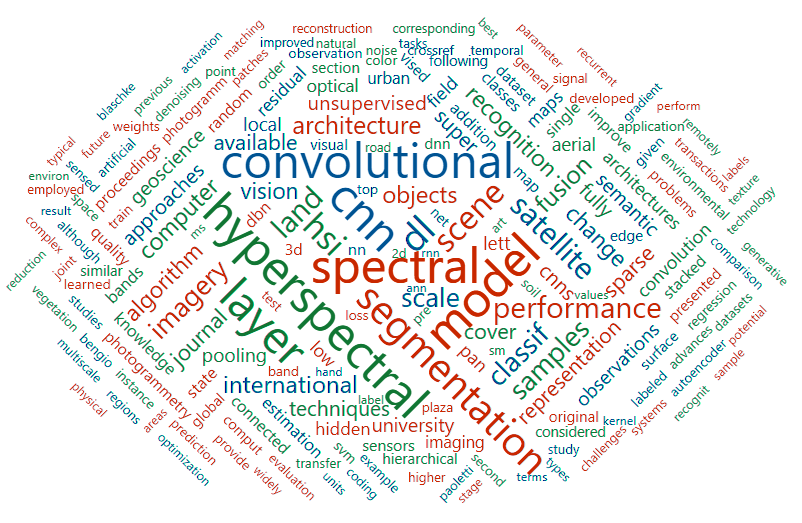}
\caption{\small \centering Word-cloud of different literature-revision papers related to the ``remote sensing'' and ``deep learning'' themes.\label{wordcloud}}
\end{figure}

The aforementioned studies were evaluated with a text processing method that returned a word cloud in which the word size denotes the frequency of the word within these papers (Fig.~\ref{wordcloud}). An interesting observation regarding this world-cloud is that the term ``UAV'' is under or not represented at all. This revision gap is a problem since UAV image data is daily produced in large amounts, and no scientific investigation appears to offer a comprehensive literature revision to assist new research on this matter. In the UAV context, there are some revision papers published in important scientific journals from the remote sensing community. As of recently, a revision-survey \citep{Bithas2019} focused on the implications of ML methods being applied to UAV image processing, but no investigation was conducted on DL algorithms for this particular issue. This is an important theme, especially since UAV platforms are more easily available to the public and DL-based methods are being tested to provide accurate mapping in highly detailed imagery.

As mentioned, UAVs offer flexibility in data collection, as flights are programmed under users' demand; they are low-cost when compared to other platforms that offer similar spatial-resolution images; produce high-level of detail in its data collection; presents dynamic data characteristics since it is possible to embed RGB, multispectral, hyperspectral, thermal and, LiDAR sensors on it; and are capable of gathering data from difficult to access places. Aside from that, sensors embedded in UAVs are known to generate data at different altitudes and point-of-views. These characteristics, alongside others, are known to produce a higher dynamic range of images than common sensing systems. This ensures that the same object is viewed from different angles, where not only their spatial and spectral information is affected, as well as form, texture, pattern, geometry, illumination, etc. This becomes a challenge for multidomain detection. As such, studies indicate that DL is the most prominent solution for dealing with these disadvantages. These studies, which most are presented in this revision paper, were conducted within a series of data criteria and evaluated DL architectures in classifying, detecting, and segmenting various objects from UAV scenes.

To the best of our knowledge, there is a literature gap related to review articles combining both ``deep learning'' and ``UAV remote sensing'' thematics. This survey is important to summarize the direction of DL applications in the remote sensing community, particularly related to UAV-imagery. The purpose of this study is to provide a brief review of DL methods and their applications to solve classification, object detection, and semantic segmentation problems in the remote sensing field. Herein, we discuss the fundamentals of DL architectures, including recent proposals. There is no intention of summarizing existing literature, but to present an examination of DL models while offering the necessary information to understand the state-of-the-art in which it encounters. Our revision is conducted highlighting traits about the UAV-based image data, their applications, sensor types, and techniques used in recent approaches in the remote sensing field. Additionally, we relate how DL models present promising results and project future perspectives of prominent paths to be explored. In short, this paper brings the following contributions:
\begin{enumerate}
\item A presentation of fundamental ideas behind the DL models, including classification, object detection, and semantic segmentation approaches; as well as the application of these concepts to attend UAV-image based mapping tasks;
\item The examination of published material in scientific sources regarding sensors types and applications, categorized in environmental, urban, and agricultural mapping contexts; 
\item The organization of publicly available datasets from previous researches, conducted with UAV-acquired data, also labeled for both object detection and segmentation tasks;
\item A description of the challenges and future perspectives of DL-based methods to be applied with UAV-based image data.
\end{enumerate}

\section{Deep Neural Networks Overview}

DNNs are based on neural networks which are composed of neurons (or units) with certain activations and parameters that transform input data (e.g., UAV remote sensing image) to outputs (e.g., land use and land cover maps) while progressively learning higher-level features \citep{Ma2019,Schmidhuber2015}. This progressive feature learning occurs, among others, on layers between the input and the output, which are referred to as hidden layers \citep{Ma2019}. DNNs are considered as a DL method in their most traditional form (i.e. with 2 or more hidden layers). Their concept, based on an Artificial Intelligence (AI) modeled after the biological neurons' connections, exists since the 1950s. But only later, with advances in computer hardware and the availability of a high number of labeled examples, its interest has resurged in major scientific fields. In the remote sensing community, the interest in DL algorithms has been gaining attention since mid 2010s decade, specifically because these algorithms achieved significant success at digital image processing tasks \citep{Ma2019,Khan2020}.

A DNN works similarly to an ANN, when as a supervised algorithm, uses a given number of input features to be trained, and that these feature observations are combined through multiple operations, where a final layer is used to return the desired prediction. Still, this explanation does not do much to highlight the differences between traditional ANNs and DNNs. LeCun et. al. \citep{Lecun2015}, the paper amongst the most cited articles in DL literature,  defines DNN as follows: ``Deep-learning methods are representation-learning methods with multiple levels of representation''. Representation-learning is a key concept in DL. It allows the DL algorithm to be fed with raw data, usually unstructured data such as images, texts, and videos, to automatically discover representations. 

\begin{figure*}[ht!]
\centering
\includegraphics[width=14cm]{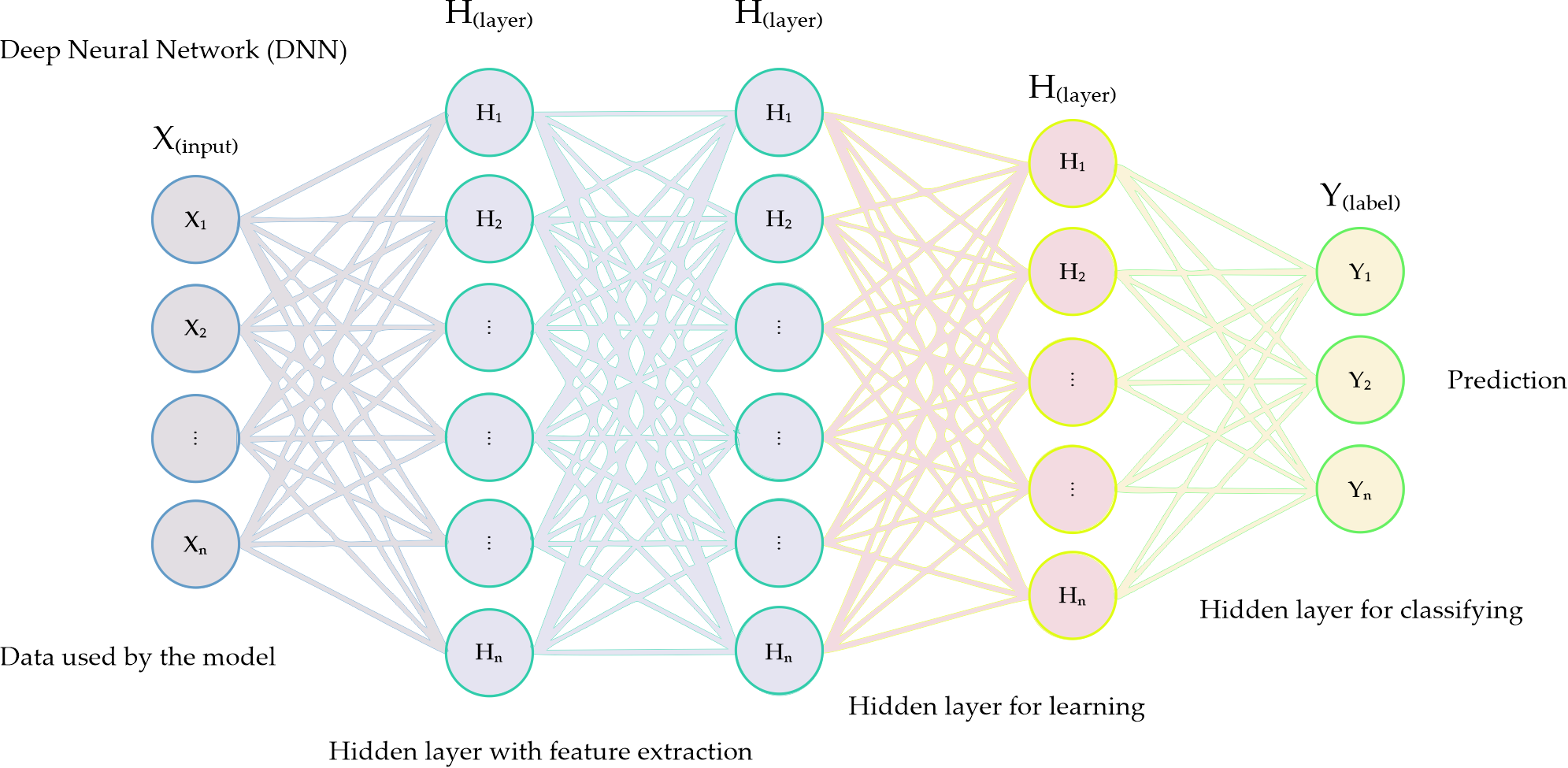}
\caption{\small \centering A DNN architecture. This is a simple example of how a DNN may be built. Here the initial layer (X\textsubscript{input}) is composed of the collected data samples. Later this data information can be extracted by hidden layers in a back-propagation manner, which is used by subsequent hidden layers to learn these features' characteristics. In the end, another layer is used with an activation function related to the given problem (classification or regression, as an example), by returning a prediction outcome (Y\textsubscript{label}).\label{dnn}}
\end{figure*}

The most common DNNs (Fig.~\ref{dnn}) are generally composed of dense layers, wherein activation functions are implemented in. Activation functions compute the weighted sum of input and biases, which is used to decide if a neuron can be activated or not \citep{Nwankpa2018}. These functions constitute decision functions that help in learning intrinsic patterns \citep{Khan2020}; i.e., they are one of the main aspects of how each neuron learns from its interaction with the other neurons. Known as a piecewise linear function type, ReLu defines the 0 valor for all negative values of X. This function is, at the time of writing, the most popular in current DNNs models. Regardless, another potential activation function recently explored is Mish, a self regularized non-monotonic activation function \citep{Khan2020}. Aside from the activation function, another important information on how a DNN works is related to its layers, such as dropout, batch-normalization, convolution, deconvolution, max-pooling, encode-decode, memory cells, and others. This layer is regularly used to solve issues with covariance-shift within feature-maps \citep{Khan2020}. The organization in which the layers are composed, as well as its parameters, is one of the main aspects of the architecture.

Multiple types of architectures were proposed in recent years to improve and optimize DNNs by implementing different kinds of layers, optimizers, loss functions, depth-level, etc. However, it is known that one of the major reasons behind DNNs' popularity today is also related to the high amount of available data to learn from it. A rule of thumb conceived among data scientists indicates that at least 5,000 labeled examples per category was recommended \citep{Goodfellow2016}. But, as of today, DNNs' proposals focused on improving these network's capacities to predict features with fewer examples than that. Some applications which are specifically oriented may benefit from it, as it reduces the amount of labor required at sample collection by human inspection. Even so, it should be noted that, although this pursuit is being conducted, multiple takes are performed by the vision computer communities and novel research includes methods for data-augmentation, self-supervising, and unsupervised learning strategies, as others. A detailed discussion of this manner is presented in \citep{Khan2020}.

\subsection{Convolutional and Recurrent Neural Networks}

\begin{figure*}[ht!]
\centering
\includegraphics[width=14cm]{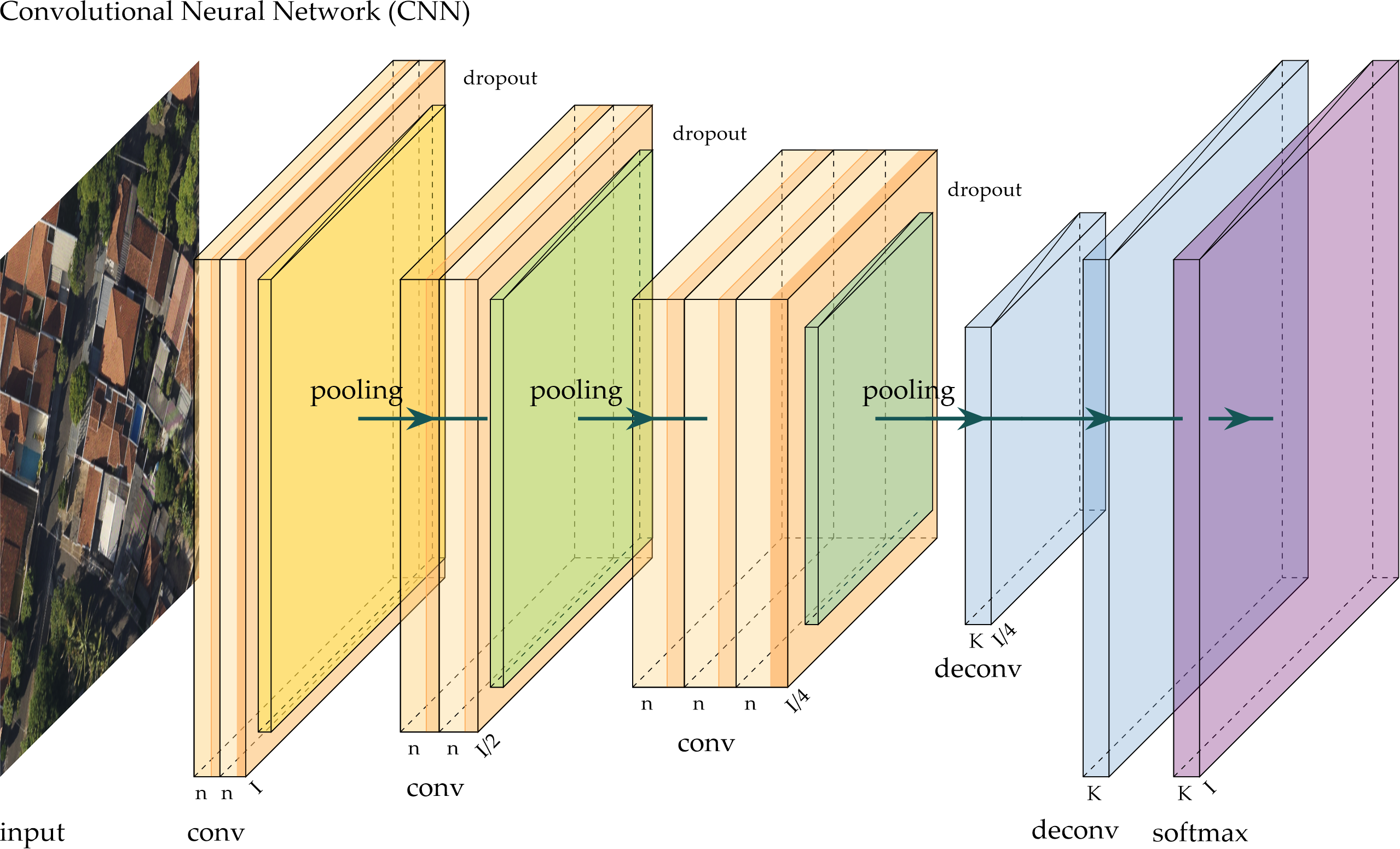}
\caption{\small \centering A CNN type of architecture with convolution and deconvolution layers. This example architecture is formed by convolutional layers, where a dropout layer is added between each conv layer, and a max-pooling layer is adopted each time the convolution window-size is decreased. By the end of it, a deconvolutional layer is used with the same size as the last convolutional, and then it uses information from the previous step to reconstruct the image with its original size. The final layer is of a softmax, where it returns the models' predictions.\label{cnn}}
\end{figure*}

A DNN can be formed by different architectures, and the complexity of the model is related to how each layer and additional computational method is implemented. Different DL architectures are proposed regularly, Convolutional Neural Networks (CNN), Recurrent Neural Networks (RNN), and Deep Belief Networks (DBN) \citep{Ball2017}, and, more recently yet, Generative Adversarial Networks (GAN) \citep{Goodfellow2016}. However, the most common DNNs in the supervised networks categories are usually classified as CNNs (Fig.~\ref{cnn}) and RNNs \citep{Khan2020}. 

\begin{figure*}[ht!]
\centering
\includegraphics[width=14cm]{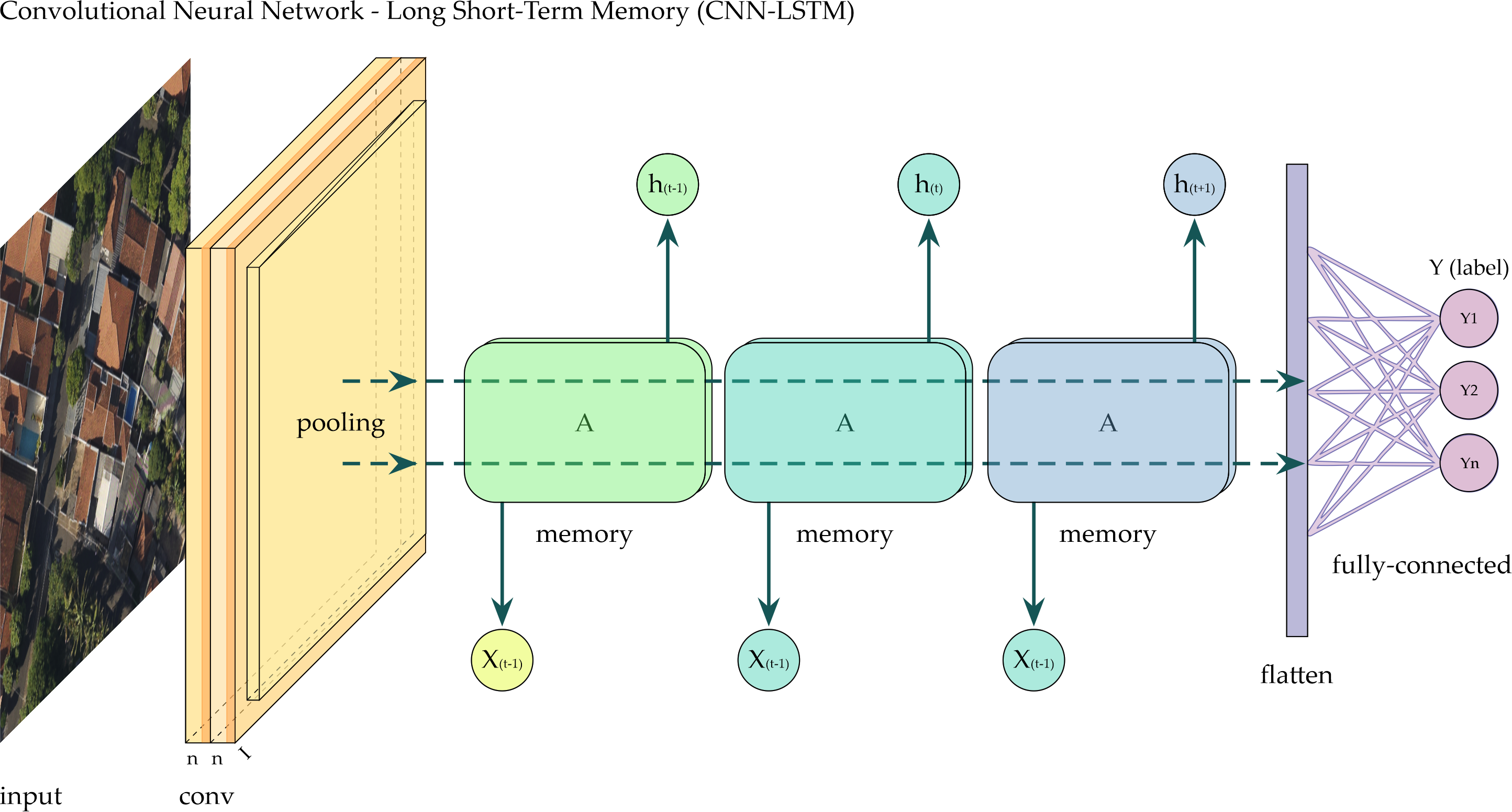}
\caption{\small \centering An example of a neural network based on the CNN-LSTM type of architecture. The input image is processed with convolutional layers, and a max-pooling layer is used to introduce the information to the LSTM. Each memory cell is updated with weights from the previous cell. After this process, one may use a flatten layer to transform the data in an arrangement to be read by a dense (fully-connected) layer, returning a classification prediction, for instance.\label{cnn_lstm}}
\end{figure*}

As a different kind of DL network structure, RNNs refer to another supervised learning model. The main idea behind implementing RNNs regards their capability of improving their learning by repetitive observations of a given phenom or object, often associated with a time-series collection. A type of RNN being currently implemented in multiple tasks is the Long Short-Term Memory (LSTM)\citep{Hochreiter1997}. In the remote sensing field, RNN models have been applied to deal with time series tasks analysis, aiming to produce, for example, land cover mapping \citep{Ienco2017,HoTongMinh2018}. For a pixel-based time series analysis aiming to discriminate classes of winter vegetation coverage using SAR Sentinel-1 \citep{HoTongMinh2018}, it was verified that RNN models outperformed classical ML approaches. A recent approach \citep{Feng2020} for accurate vegetation mapping  combined multiscale CNN to extract spatial features from UAV-RGB imagery and then fed an attention-based RNN to establish the sequential dependency between multitemporal features. The aggregated spatial-temporal features are used to predict the vegetable category. Such examples with remote sensing data demonstrate the potential in which RNNs are being used. Also, one prominent type of architecture is the CNN-LSTM method (Fig.~\ref{cnn_lstm}). This network uses convolutional layers to extract important features from the given input image and feed the LSTM. Although few studies implemented this type of network, it should be noted that it serves specific purposes, and its usage, for example, can be valued for multitemporal applications.

As aforementioned, other types of neural networks, aside from CNNs and RNNs, are currently being proposed to also deal with an image type of data. GANs are amongst the most innovative unsupervised DL models. GANs are composed of two networks: generative and discriminative, that contest between themselves. The generative network is responsible for extracting features from a particular data distribution of interest, like images, while the discriminative network distinguishes between real (reference or ground truth data) and those data generated by the generative part of GANs (fake data) \citep{Goodfellow2014, Ma2019}. Recently approaches in the image processing context like the classification of remote sensing images \citep{Lin2017} and image-to-image translation problems solution \citep{Isola2018} adopted GANs as DL model, obtaining successful results.

\begin{figure*}[ht!]
\centering
\includegraphics[width=16.5cm]{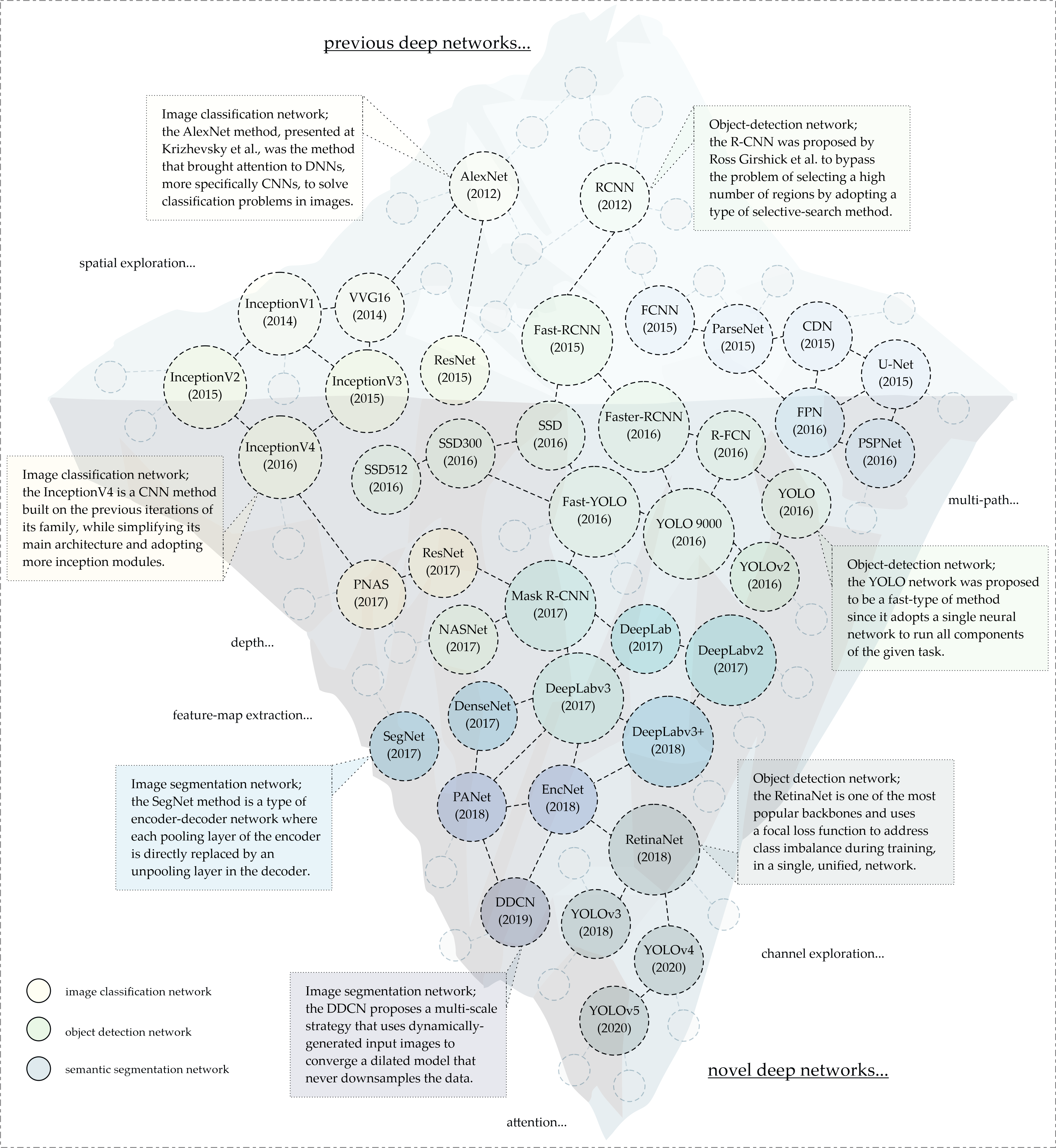}
\caption{\small \centering A DL time-series indicating some popular architectures implemented in image classification (yellowish color), object detection (greenish color), and segmentation (bluish color). These networks often intertwine, and many adaptations have been proposed for them. Although it may appear that most of the DL methods were developed during 2015-2017 annuals, it is important to note that, as some, novel deep networks use most of the already developed methods as backbones, or accompanied from other types of architectures, mainly used as the feature extraction part of a much more complex structure.\label{iceberg}}
\end{figure*}

In short, several DNNs are constantly developed, in both scientific and/or image competition platforms, to surpass existing methods. However, as each year passes, some of these neural networks are often mentioned, remembered, or even improved by novel approaches. A summary of well-known DL methods built in recent years is presented in Fig.~\ref{iceberg}. A detailed take on this, which we recommend to anyone interested, is found in Khan et al. \citep{Khan2020}. Alongside the creations and developments of these and others, researchers observed that higher depth channel exploration, and, as of recently proposed, attention-based feature extraction neural networks, are regarded as some of the most prominent approaches for DL. Initially, most of the proposed supervised DNNs, like CNN and RNN, or CNN-LSTM models, were created to perform and deal with specific issues. Often, these approaches can be grouped into classification tasks, like scene-wise classification, object detection, semantic and instance segmentation (pixel-wise), and regression tasks.

\subsection{Classification and Regression Approaches}

When considering remote sensing data processed with DL-based algorithms, the following tasks can be highlighted: scene-wise classification, semantic and instance segmentation, and object detection. Scene-wise classification involves assigning a class label to each image (or patch), while the object detection task aims to draw bounding boxes around objects in an image (or patch) and labeling each of them according to the class label. Object detection can be considered a more challenging task since it requires to locate the objects in the image and then perform their classification. Another manner to detect objects in an image, instead of drawing bounding boxes, is to draw regions or structures around the boundary of objects, i.e., distinguish the class of the object at the pixel level. This task is known as semantic segmentation. However, in semantic segmentation, it is not possible to distinguish multiple objects of the same category, as each pixel receives one class label \citep{Wu2020}. To overcome this drawback, a task that combines semantic segmentation and object detection named instance segmentation was proposed to detect multiple objects in pixel-level masks and labeling each mask with a class label \citep{thoma2016survey, chen2016semantic}. The instance segmentation, however, consists of a method that, while classifying the image with this pixel-wise approach, is able to individualize objects \citep{Sharma2020}.

To produce a deep regression approach, the model needs to be adapted so that the last fully-connected layer of the architecture is changed to deal with a regression problem instead of a common classification one. With this adaptation, continuous values are estimated, differently from classification tasks. In comparison to classification, the regression task using DL is not often used; however, recent publications have shown its potential in remote sensing applications. One approach \citep{Lathuiliere2020} performed a comprehensive analysis of deep regression methods and pointed out that well-known fine-tuned networks, like VGG-16 \citep{VGG15} and ResNet-50 \citep{He2016}, can provide interesting results. These methods, however, are normally developed for specific applications, which is a drawback for general-purpose solutions. Another important point is that depending on the application, not always deep regression succeeds. A strategy is to discretize the output space and consider it as a classification solution. For UAV remote sensing applications, the strategy of using well-known networks is in general adopted. Not only VGG-16 and ResNet-50, as investigated by \citep{Lathuiliere2020}, but also other networks including AlexNet \citep{Krizhevsky2012} and VGG-11 have been used. An important issue that could be investigated in future research, depending on the application, is the optimizer. Algorithms with adaptive learning rates such as AdaGrad, RMSProp, AdaDelta (an extension of AdaGrad), and Adam are among the commonly used.

\subsubsection{Scene-Wise Classification, Object Detection, and Segmentation}

\begin{figure*}[ht!]
\centering
\includegraphics[width=11.5cm]{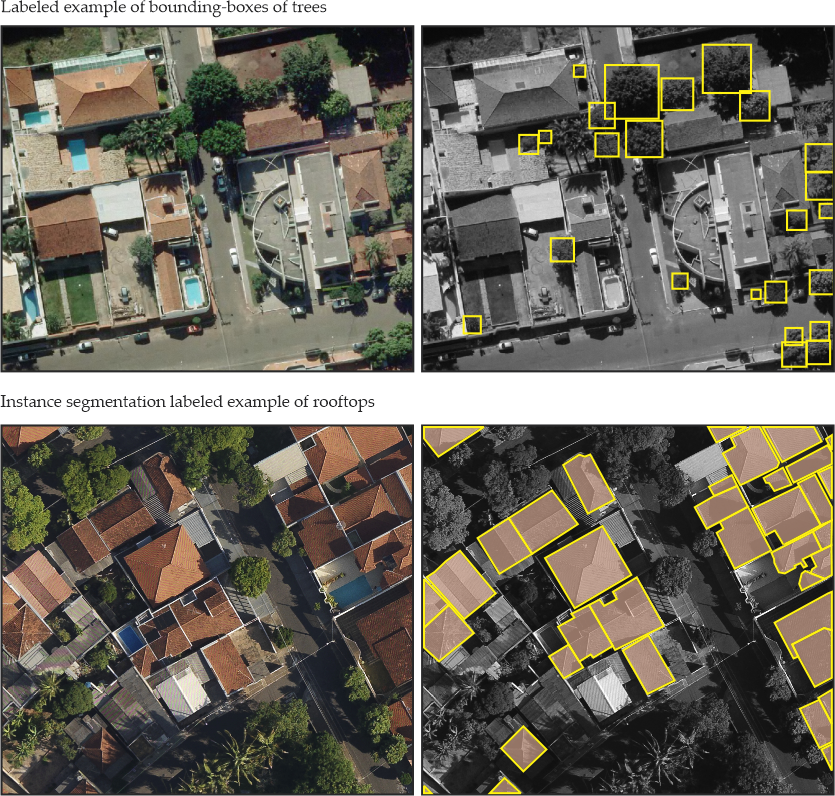}
\caption{\small \centering Labeled examples. The first-row consists of a bounding-box type of object detection approach label-example to identify individual tree-species in an urban environment. The second-row is a labeled-example of instance segmentation to detect rooftops in the same environment.\label{label}}
\end{figure*}

Scene-wise classification or scene recognition refers to methods that associate a label/theme for one image (or patch) based on numerous images, such as in agricultural scenes, beach scenes, urban scenes, and others \citep{Zou2015, Ma2019}. Basic DNNs methods were developed for this task, and they are among the most common networks for traditional image recognition tasks. In remote sensing applications, scene-wise classification is not usually applied. Instead, most applications benefit more from object detection and pixel-wise semantic segmentation approaches. For scene-wise classification, the method needs only the annotation of the class label of the image, while other tasks like object detection method needs a drawn of a bounding box for all objects in an image, which makes it more costly to build labeled datasets. For instance or semantic segmentation, the specialist (i.e., the person who performs the annotation or object labeling) needs to draw a mask involving each pixel of the object, which needs more attention and precision in the annotation task, reducing, even more, the availability of datasets. Fig.~\ref{label} shows the examples of both annotation approaches (object detection and instance segmentation).

Object detection methods can be described into two mainstream categories: one-stage detectors (or regression-based methods) and two-stage detectors (or region proposal-based methods) \citep{Zhao2019,Liu2019,Wu2020}. The usual two-stage object detection pipeline is to generate region proposals (candidate rectangular bounding boxes) on the feature map. It then classifies each one into an object class label and refines the proposals with a bounding box regression. A widely used strategy in the literature to generate proposals was proposed with the Faster-RCNN algorithm with the Region Proposal Network (RPN) \citep{Zhao2019}. Other state-of-the-art representatives of such algorithms are Cascade-RCNN \citep{CascadeRCNN}, Trident-Net \citep{Tridentnet}, Grid-RCNN \citep{GridRCNN}, Dynamic-RCNN \citep{DynamicRCNN}, DetectoRS \citep{Detectors}. As for one-stage detectors, they directly make a classification and detect the location of objects without a region proposal classification step. This reduced component achieves a high detection speed for the models but tends to reduce the accuracy of the results. These are known as region-free detectors since they typically use cell grid strategies to divide the image and predict the class label of each one. Besides that, some detectors may serve for both one-stage and two-stage categories.

Object detection-based methods can be described in three components: a) backbone, which is responsible to extract semantic features from images; b) the neck, which is an intermediate component between the backbone and the head components, used to enrich the features obtained by the backbone, and; c) head component, which performs the detection and classification of the bounding boxes. 

The backbone is a CNN that receives as input an image and outputs a feature map that describes the image with semantically features. In the DL, the state-of-the-art is composed of the following backbones: VGG \citep{VGG15}, ResNet \citep{Resnet16}, ResNeXt \citep{Resnext17}, HRNet \citep{Hrnet20}, RegNet \citep{Regnet20}, Res2Net \citep{Res2net21}, and ResNesT \citep{Resnest20}. The neck component combines in several scales low-resolution and semantically strong features, capable of detecting large objects, with high-resolution and semantically weak features, capable of detecting small objects, which is done with the lateral and top-down connections of the convolutional layers of the Feature Pyramid Network (FPN) \citep{Fpn17}, and its variants like PAFPN \citep{Pafpn18} and NAS-FPN \citep{Nasfpn19}. Although FPN was originally designed to be a two-stage method,  the methods' purpose was a manner to use the FPN on single-stage detectors by removing RPN and adding a classification subnet and a bounding box regression subnet. The head component is responsible for the detection of the objects with the softmax classification layer, which produces probabilities for all classes and a regression layer to predict the relative offset of the bounding box positions with the ground truth.

Despite the differences in object detectors (one or two-stage), their universal problem consists of dealing with a large gap between positive samples (foreground) and negative samples (background) during training, i.e class imbalance problem that can deteriorate the accuracy results \citep{Chen2020}. In these detectors, the candidate bounding boxes can be represented into two main classes: positive samples, which are bounding boxes that match with the ground-truth, according to a metric; and negative samples, which do not match with the ground-truth. In this sense, a non-max suppression filter can be used to refine these dense candidates by removing overlaps to the most promising ones. The Libra-RCNN \citep{Pang2019}, ATSS \citep{Atss19}, Guided Anchoring \citep{Ga19}, FSAF \citep{Zhu2019}, PAA \citep{Paa20}, GFL \citep{Gfl20}, PISA \citep{Pisa20} and VFNet \citep{Vfnet20} detectors explore different sampling strategies and new loss metrics to improve the quality of selected positive samples and reduce the weight of the large negative samples.

Another theme explored in the DL literature is the strategy of encoding the bounding boxes, which influences the accuracy of the one-stage detectors as they do not use region proposal networks \citep{Vfnet20}. In this report \citep{Vfnet20}, the authors represent the bounding boxes like a set of representatives or key-points and find the farthest top, bottom, left, and right points. CenterNet \citep{Duan2019} detects the object center point instead of using bounding boxes, while CornerNet \citep{Law2020} estimates the top-left corner and the bottom-right corner of the objects. SABL \citep{Sabl20} uses a chunk based strategy to discretize horizontally and vertically the image and estimate the offset of each side (bottom, up, left, and right). The VFNet \citep{Vfnet20} method proposes a loss function and a star-shaped bounding box (described by nine sampling points) to improve the location of objects.

Regarding semantic segmentation and instance segmentation approaches, they are generally defined as a pixel-level classification problem \citep{Segsurvey21}. The main difference between semantic and instance is that the former one is capable to identify pixels belonging to one class but can not distinguish objects of the same class in the image. However, instance segmentation approaches can not distinguish overlapping of different objects, since they are concerned with identifying objects separately. For example, it may be problematic to identify in an aerial urban image the location of the cars, trucks, motorcycle, and the asphalt pavement which consists of the background or region in which the other objects are located. To unify these two approaches, a method was recently proposed in \citep{PanSeg19}, named panoptic segmentation. With panoptic segmentation, the pixels that are contained in uncountable regions (e.g. background) receive a specific value indicating it.

Considering the success of the RPN method for object detection, some variants of Faster R-CNN were considered to instance segmentation as Mask R-CNN \citep{Maskrcnn17}, which in parallel to bounding box regression branch add a new branch to predict the mask of the objects (mask generation). The Cascade Mask R-CNN \citep{Cascademask19} and HTC \citep{Htc19} extend Mask R-CNN to refine in a cascade manner the object localization and mask estimation. The PointRend \citep{Pointrend20} is a point-based method that reformulates the mask generation branch as a rendering problem to iteratively select points around the contour of the object. Regarding semantic segmentation, methods like U-Net \citep{Ronneberger2015}, SegNet \citep{Badrinarayanan2017}, DeepLabV3+ \citep{Chen2018}, and Deep Dual-domain Convolutional Neural Network (DDCN) \citep{Nogueira2019} have also been regularly used and adapted for recent remote sensing investigations \citep{Nogueira2020}. Another important remote sensing approach that is been currently investigated is the segmentation of objects considering sparse annotations \citep{hua2021sparse}. Still, as of today, the CGnet \citep{Cgnet20} and DLNet \citep{dnl20} are considered the state-of-art methods for semantic segmentation.

\section{Deep Learning in UAV Imagery }

To identify works related to DL in UAV remote sensing applications, we performed a search in the Web of Science (WOS) and Google Scholar databases. WOS is one of the most respected scientific databases and hosts a high number of scientific journals and publications. We conducted a search using the following string in the WOS: (``TS = ((deep learning OR CNN OR convolutional neural network) AND (UAV OR unmanned aerial vehicle OR drone OR RPAS) AND (remote sensing OR photogrammetry)) AND LANGUAGE: (English) AND Types of Document: (Article OR Book OR Book Chapter OR Book Review OR Letter OR Proceedings Paper OR Review); Indexes=SCI-EXPANDED, SSCI, A\%HCI, CPCI-S, CPCI-SSH, ESCI. Stipulated-time=every-years.''). We considered DL, but added CNN, as it is one of the main DL-based architectures used in remote sensing applications \citep{Ma2019}. As such, published materials that use these terms in their titles, abstracts or keywords were investigated and included. For such reasons, we opted for this string to achieve a generalist investigation.

\begin{figure*}[ht!]
\centering
\includegraphics[width=14.5cm]{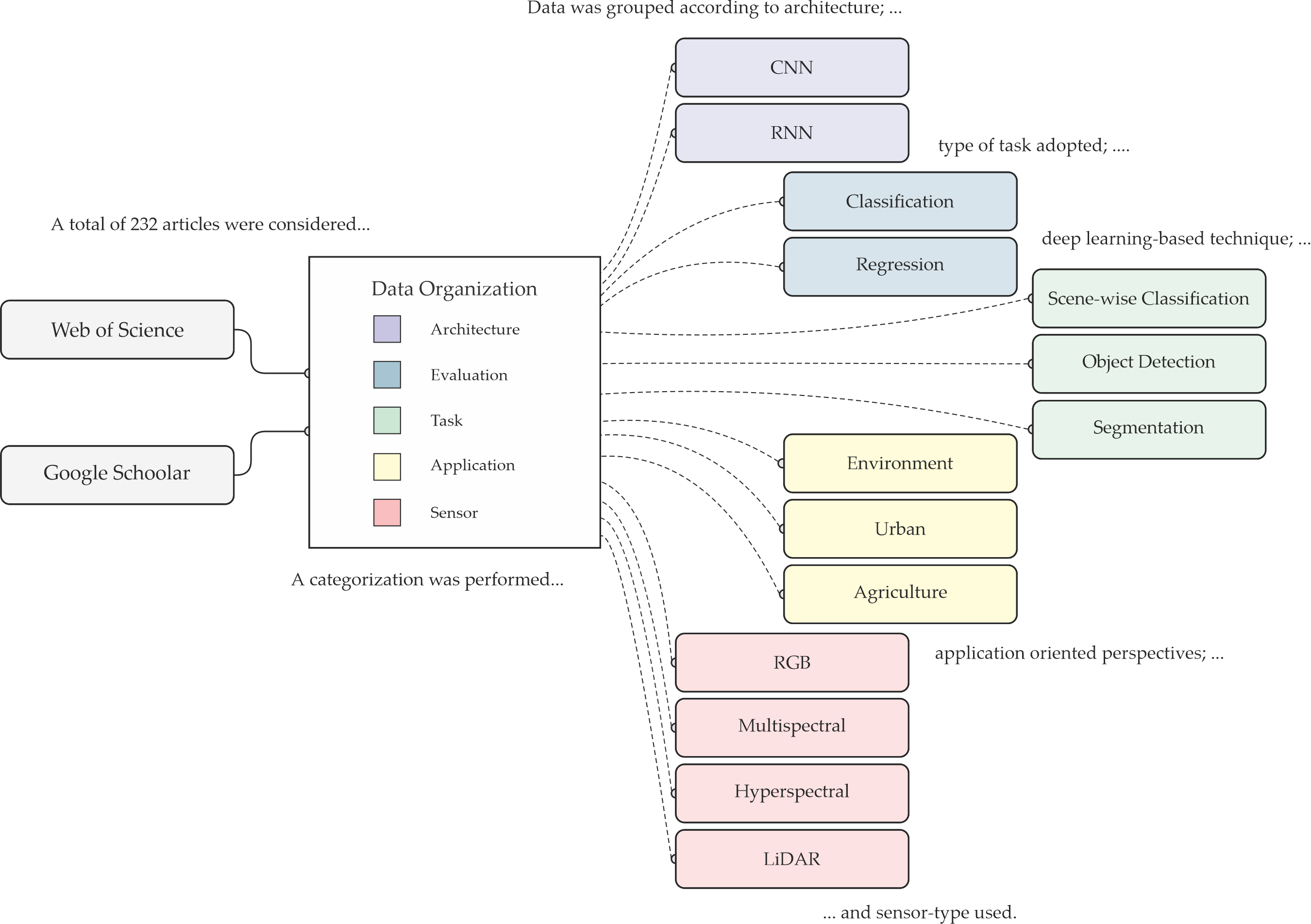}
\caption{\small \centering The schematic procedure adopted to organize the revised material according to their respective categories as proposed in this review.\label{flow}}
\end{figure*}

We filtered the results to consider only papers that implemented approaches with UAV-based systems. A total of 190 papers were found in the WOS database, where 136 were articles, 46 proceedings, and 10 reviews. An additional search was conducted in the Google Scholar database to identify works not detected in the WOS. We adopted the same combination of keywords in this search. We performed a detailed evaluation of its results and selected only those that, although from respected journals, were not encountered in the WOS search. This resulted in a total of 34 articles, 16 proceedings, and 8 reviews. The entire dataset was composed of 232 articles + proceedings and 18 reviews from scientific journals indexed in those bases. These papers were then organized and revised. Fig.~\ref{flow} demonstrates the main steps to map this research. The encountered publications were registered only in the last five years (from 2016 to 2021), which indicates how recent UAV-based approaches integrated with DL methods are in the scientific journals.

The review articles gathered at those bases were separated and mostly used in the cloud text analysis of Fig.~\ref{wordcloud}, while the remaining papers (articles and proceedings) were organized according to their category. A total of 283.785 words were analyzed for the word-cloud, as we removed words with less than 5\% occurrences to cut lesser-used words unrelated to the theme, and higher than 95\% occurrences to remove plain and simple words frequently used in the English language. The published articles and proceedings were divided in terms of DL-based networks (classification: scene-wise classification, segmentation, and object detection and; regression), sensor types (RGB, multispectral, hyperspectral, and LiDAR); and; applications (environmental, urban, and agricultural context). We also provided, in a subsequent section, datasets from previously conducted research for further investigation by novel studies. These datasets were organized and their characteristics were also summarized accordingly.

\begin{figure*}[ht!]
\centering
\includegraphics[width=14.5cm]{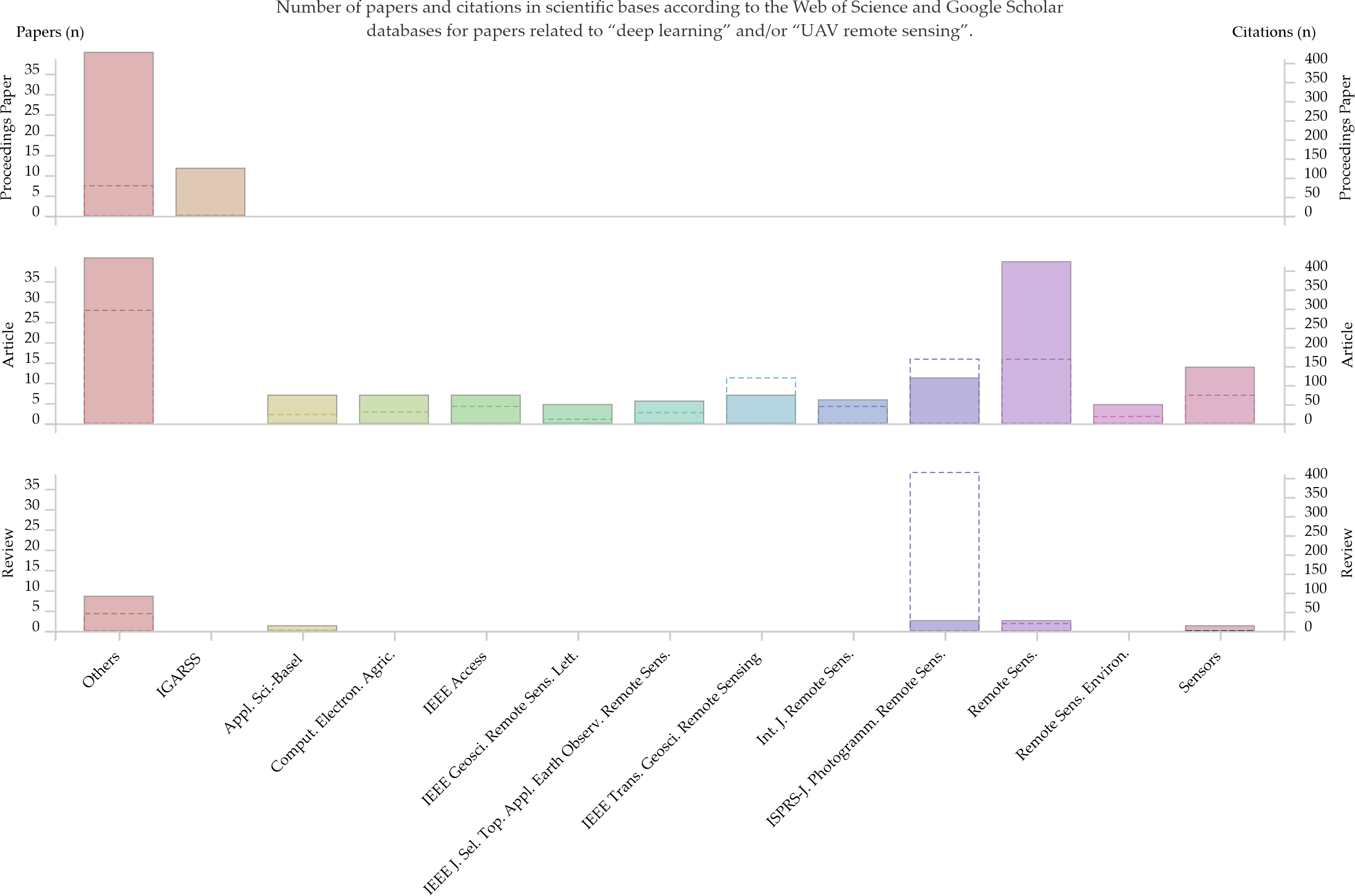}
\caption{\small \centering The distribution of the evaluated scientific material according to data gathered at Web of Science (WOS) and Google Scholar databases. The y-axis on the left represents the number (n) of published papers, illustrated by solid-colored boxes. The y-axis on the right represents the number of citations that these publications, according to peer-review scientific journals, received since their publication, illustrated by dashed-lines of the same color to its corresponding solid-colored box.\label{paper}}
\end{figure*}

Most of our research was composed of publications from peer-review publishers in the area of remote sensing journals (Fig.~\ref{paper}). Even though the review articles encountered in the WoS and Google Scholar databases do mention, to some extent, UAV-based applications, none of them were dedicated to it. Towards the end of our paper, we examined state-of-the-art approaches, like real-time processing, data dimensionality reduction, domain adaptation, attention-based mechanisms, few-shot learning, open-set, semi-supervised and unsupervised learning, and others. This information provided an overview of the future opportunities and perspectives on DL methods applied in UAV-based images, where we discuss the implications and challenges of novel approaches.

The 232 papers (articles + proceedings) were investigated through a quantitative perspective, where we evaluated the number of occurrences per journal, the number of citations, year of publication, and location of the conducted applications according to country. We also prepared and organized a sampling portion in relation to the corresponding categories, as previously explained, identifying characteristics like architecture used, evaluation metric approach, task conducted, and type of sensor and mapping context objectives. After evaluating it, we adopted a qualitative approach by revising and presenting some of the applications conducted within the papers (UAV + DL) encountered in the scientific databases, summarizing the most prominent ones. This narrative over these applications was separated accordingly to the respective categories related to the mapping context (environmental, urban, and agricultural). Later on, when presenting future perspectives and current trends in DL, we mentioned some of these papers alongside other investigations proposed at computer vision scientific journals that could be potentially used for remote sensing and UAV-based applications.

\subsection{Sensors and Applications Worldwide}

\begin{figure*}[ht!]
\centering
\includegraphics[width=14.5cm]{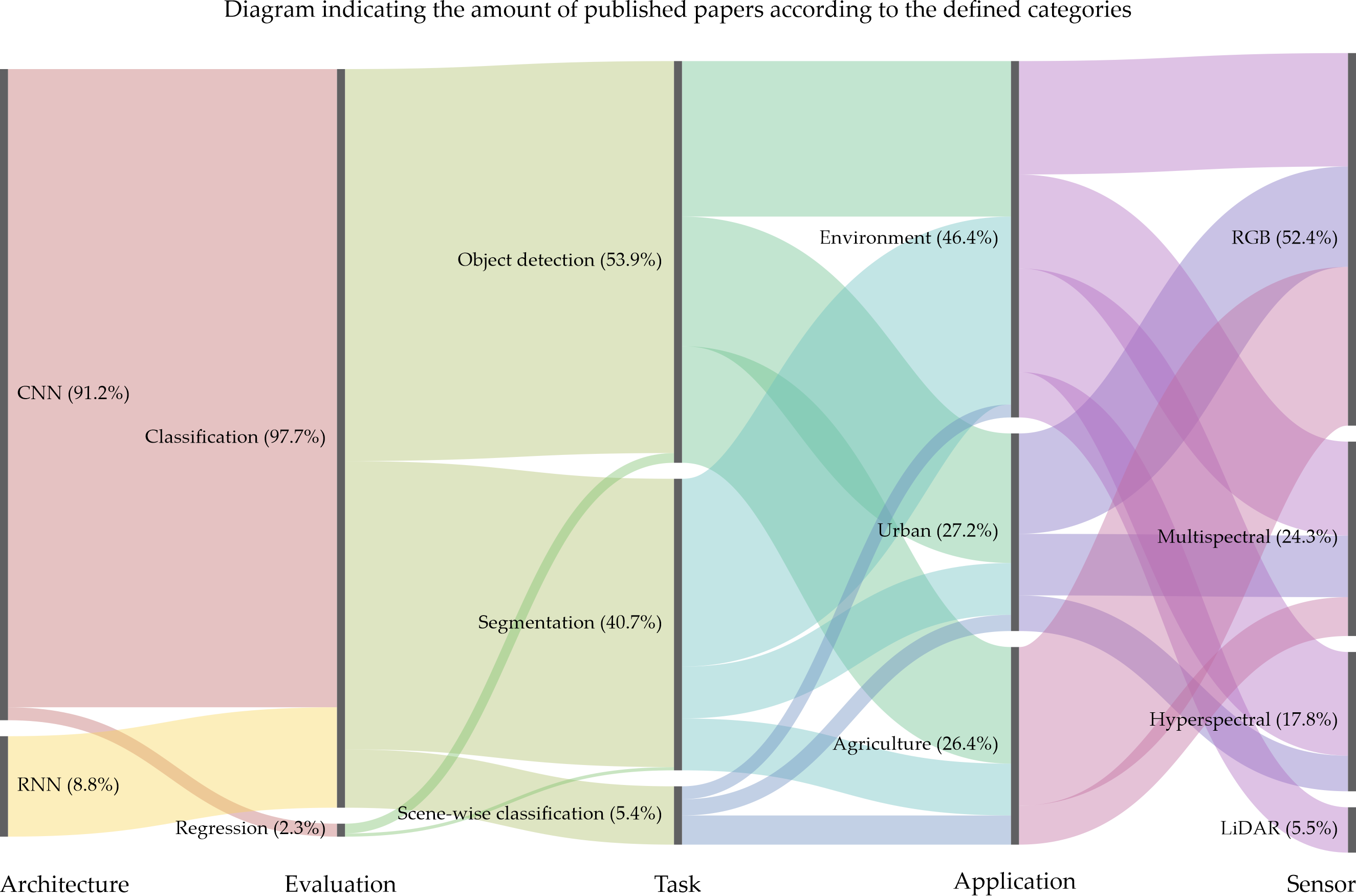}
\caption{\small \centering Diagram describing proceedings and articles according to the defined categories using WOS and Google Scholar datasets.\label{diagram}}
\end{figure*}

In the UAV-based imagery context, several applications were beneficiated from DL approaches. As these networks' usability is increasing throughout different remote sensing areas, researchers are also experimenting with their capability in substituting laborious-human tasks, as well as improving traditional measurements performed by shallow learning or conventional statistical methods. As of recently, several articles and proceedings were published in renowned scientific journals. In general terms, the articles collected at the scientific databases demonstrated a pattern related to its architecture (CNN or RNN), evaluation (classification or regression) approach (object detection, segmentation, or scene-wise classification), type of sensor (RGB, multispectral, hyperspectral or LiDAR) and mapping context (environmental, urban, or agricultural). These patterns can be viewed on a diagram (Fig.~\ref{diagram}). The following observations can be extracted from this graphic: 
\begin{enumerate}
\item The majority of networks in UAV-based applications still rely mostly on CNNs; 
\item Even though object detection is the highest type of approach, there has been a lot of segmentation approaches in recent years; 
\item Most of the used sensors are RGB, followed by multispectral, hyperspectral, and LiDAR, and;
\item There is an interesting amount of papers published within the environmental context, with forest-type related applications being the most common approach in this category, while both urban and agricultural categories were almost evenly distributed among opted approaches.
\end{enumerate}

The majority of papers published on UAV-based applications implemented a type of CNN (91.2\%). Most of these articles used established architectures (Fig.~\ref{iceberg}) and a small portion proposed their models and compared them against the state-of-the-art networks. In reality, this comparison appears to be a crucial concern regarding recent publications, since it is necessary to ascertain the performance of the proposed method in relation to well-known DL-based models. Still, the popularity of CNNs architecture in remote sensing images is not new, mainly because of reasons already stated in the previous sections. Besides that, even though presented in a small number of articles, RNNs (8.8\%), mostly composed of CNN-LSTM architectures, are an emerging trend in this area and appear to be the focus of novel proposals. As UAV systems are capable of operating mostly according to the users' own desires (i.e., can acquire images from multiple dates in a more personalized manner), the same object is viewed through a type of time-progression approach. This is beneficial for many applications that include monitoring of stationary objects, like rivers, vegetation, or terrain slopes, for example.

Although classification (97.7\%) tasks are the most common evaluation metrics implemented in these papers, regression (2.3\%) is an important estimate and may be useful in future applications. The usage of regression metrics in remote sensing applications is worth it simply because it enables the estimation of continuous data. Applications that could benefit from regression analysis are present in environmental, urban, and agricultural contexts, as in many others, and it is useful to return predictions on measured variables. Classification, on the other hand, is more of a common ground for remote sensing approaches and it is implemented in every major task (object detection; pixel-wise semantic segmentation and scene-wise classification).

The aforementioned DL-based architectures were majorly applied in object detection (53.9\%) and image segmentation (40.7\%) problems, while (scene-wise) classification (5.4\%) were the least common. This preference for object detection may be related to UAV-based data, specifically, since the high amount of detail of an object provided by the spatial resolution of the images is both an advantage and a challenge. It is an advantage because it increases the number of objects to be detected on the surface (thus, more labeled examples), and it is a challenge because it difficulties both the recognition and segmentation of these objects (higher detail implies more features to be extracted and analyzed). Classification (scene-wise), on the other hand, is not as common in remote sensing applications, and image segmentation is often preferred in some applications since assigning a class to each pixel of the image has more benefits for this type of analysis than rather only identifying a scene.

Following it, there is an interesting distribution pattern related to the application context. The data indicated that most of the applications were conducted in the environmental context (46.6\%). This context includes approaches that aim to, in a sense, deal with detection and classification tasks on land use and change, environmental hazards and disasters, erosion estimates, wild-life detection, forest tree inventory, monitoring difficult to access regions, as others. Urban and agricultural categories (both 27.2\% and 26.4\%, respectively) were associated with car and traffic detection, buildings, street, and rooftop extraction, as well as plant counting, plantation-row detection, weed infestation identification, and others. Interestingly, all of the LiDAR data applications were related to environmental mapping, while RGB images were mostly used for urban, followed by the agricultural context. Multispectral and hyperspectral data, however, were less implemented in the urban context in comparison against the other categories. As these categories benefit differently from DL-based methods, a more detailed intake is needed to understand its problems, challenges, and achievements. In the following subsections, we explain these issues and advances while citing some suitable examples from within our search database.

Lastly, another important observation to be made regarding the categorization division used here is that there is a visible dichotomy between the types of sensor used. Most of the published papers in this area evaluating the performance of DL-based networks with RGB sensors (52.4\%). This was, respectively, followed by multispectral (24.3\%), hyperspectral (17.8\%), and LiDAR (5.5\%). The preference for RGB sensors in UAV-based systems may be associated with their low-cost and high market availability. As such, published articles may reflect on this, since it is a viable option for practical reasons when considering the replicability of the method. It should be noted that the number of labeled examples in public databases are mostly RGB, which helps improvements and investigation with this type of data. Moreover, data obtained from multispectral, hyperspectral, and LiDAR sensors are used in more specific applications, which contributes to this division. 

Most of the object detection applications went on RGB types of data, while segmentation problems were dealt with both RGB, multispectral, hyperspectral, and LiDAR data. A possible explanation for this is that object detection often relies on the spatial, texture, pattern, and shape characteristics of the object in the image, as segmentation approaches are a diverse type of applications, which benefit from the amount of spectral and terrain information provided by these sensors. In object detection, DL-based methods may have potentialized the usage of RGB images, since simpler and traditional methods need additional spectral information to perform it. Also, apart from the spectral information, LiDAR, for example, offers important features of the objects for the networks to learn and refine the edges around them, specifically where their patterns are similar. Regardless, many of these approaches are related to the available equipment and nature of the application itself, so it is difficult to pinpoint a specific reason. 

\subsection{Environmental Mapping}

Environmental approaches with DNNs-based methods hold the most diverse applications with remote sensing data, including UAV-imagery. These applications adopt different sensors simply because of their divergent nature. To map natural habits and their characteristics, studies often relied on methods and procedures specifically related to its goals, and no ``universal'' approach could be proposed nor discovered. However, although DL-based methods have not reached this type of ``universal'' approach, they are changing some skepticism by being successfully implemented in the most unique scenarios. Although UAV-based practices still offer some challenges to both classification and regression tasks, DNNs methods are proving to be generally capable of performing such tasks. Regardless, there is still much to be explored. 

Several environmental practices could potentially benefit from deep networks like CNNs and RNNs. For example, monitoring and counting wild-life \citep{Barbedo2020,Hou2020,Sundaram2020}, detecting and classifying vegetation from grasslands and heavily-forested areas \citep{Horning2020,Hamdi2019}, recognizing fire and smoke signals \citep{Larsen2020,Zhang2019b}, analyzing land use, land cover, and terrain changes, which are often implemented into environmental planning and decision-making models \citep{Kussul2017, Zhang2020}, predicting and measuring environmental hazards \citep{VanDao2020,Bui2020}, among others. What follows is a brief description of recent material published in the remote sensing scientific journals that aimed to solve some of these problems by integrating data from UAV embedded sensors with DL-based methods.

One of the most common approaches related to environmental remote sensing applications regards land use, land cover, and other types of terrain analysis. A recent study \citep{Giang2020} applied semantic segmentation networks to map land use over a mining extraction area. Another one, \citep{Al-Najjar2019}, combined information from a Digital Surface Model (DSM) with UAV-based RGB images and applied a type of feature fusion as input for a CNN model. To map coastal regions, an approach \citep{Buscombe2018}, with RGB data registered at multiple scales, used a CNN in combination with a graphical method named conditional random field (CRF). Another research \citep{Park2020}, with hyperspectral images in combination between 2D and 3D convolutional layers, was developed to determine the discrepancy of land cover in the assigned land category of cadastral map parcels. 

With a semantic segmentation approach, road extraction by a CNN was demonstrated in another investigation \citep{Li2019b}. Another study \citep{Gevaert2020} investigated the performance of a FCN to monitor household upgrading in unplanned settlements. Terrain analysis is a diversified topic in any type of cartographic scale, but for UAV-based images, in which most data acquisitions are composed by a high-level of detail, DL-based methods are resulting in important discoveries, demonstrating the feasibility of these methods to perform this task. Still, although these studies are proving this feasibility, especially in comparison with other methods, novel research should focus on evaluating the performance of deep networks regarding their domain adaptation, as well as its generalization ability, like using data in different spatial resolutions, multitemporal imagery, etc.

The detection, evaluation, and prediction of flooded areas represents another type of investigation with datasets provided by UAV-embedded sensors. A study \citep{Gebrehiwot2019} demonstrated the importance of CNNs for the segmentation of flooded regions, where the network was able to separate water from other targets like buildings, vegetation, and roads. One potential application that could be conducted with UAV-based data, but still needs to be further explored, is mapping and predicting regions of possible flooding with a multitemporal analysis, for example. This, as well as many other possibilities related to flooding, water-bodies, and river courses \citep{Carbonneau2020}, could be investigated with DL-based approaches. 

For river analysis, an investigation \citep{Zhang2020b} used a CNN architecture for image segmentation by fusing both the positional and channel-wise attentive features to assist in river ice monitoring. Another study \citep{Jakovljevic2019} compared LiDAR data with point cloud generated by UAV mapping and demonstrated an interesting approach to DL-based methods applications for point cloud classification and a rapid Digital Elevation Model (DEM) generation for flood risk mapping. One type of application with CNN in UAV data involved measuring hailstones in open areas \citep{Soderholm2020}. For this approach, image segmentation was used in RGB images and returned the maximum dimension and intermediate dimension of the hailstones. Lastly, on this topic, a comparison \citep{Ichim2020} with CNNs and GANs to segment both river and vegetation areas demonstrated that a type of ``fusion'' between these networks using a global classifier had an advantage of increasing the efficiency of the segmentation.

UAV-based forest mapping and monitoring is also an emerging approach that has been gaining the attention of the scientific community and, at some level, governmental bodies. Forest areas often pose difficulties for precise monitoring and investigation, since they can be hard to access and may be dangerous to some extent. In this aspect, images taken from UAV embedded sensors can be used to identify single tree-species in forested environments and compose an inventory. From the papers gathered, multiple types of sensors, RGB, both multi and hyperspectral, and LiDAR, were used for this approach. An application investigated the performance of a 3D-CNN method to classify tree species in a boreal forest, focusing on pine, spruce, and birch trees, with a combination between RGB and hyperspectral data \citep{Nezami2020}. 

Single-tree detection and species classification by CNNs were also investigated in \citep{Ferreira2020} in which three types of palm-trees in the Amazon forest, considered important for its population and native communities, were mapped with this type of approach. Another example \citep{Hu2020} includes the implementation of a Deep Convolutional Generative Adversarial Network (DCGAN) to discriminate between health diseased pinus-trees in a heavily-dense forested park area. Another recent investigation \citep{Miyoshi2020} proposed a novel DL method to identify single-tree species in highly-dense areas with UAV- hyperspectral imagery. These and other scientific studies demonstrate how well DL-based methods can deal with such environments.

Although the majority of approaches encountered at the databases of this category relate to tree-species mapping, UAV-acquired data were used for other applications in these natural environments. A recent study \citep{Zhang2020c} proposed a method based on semantic segmentation and scene-wise classification of plants in UAV-based imagery. The method bases itself on a CNN that classifies individual plants by increasing the image scale while integrating features learned from small scales. This approach is an important intake in multi-scale information fusion. Also related to vegetation identification, multiple CNNs architectures were investigated in \citep{Hamylton2020} to detect between plants and non-type of plants with UAV-based RGB images achieving interesting performance.

Another application aside from vegetation mapping involves wild-life identification. Animal monitoring in open spaces and grasslands is also something that received attention as DL-based object detection and semantic segmentation methods are providing interesting outcomes. A paper by \citep{Kellenberger2020} covers this topic and discusses, with practical examples, how CNNs may be used in conjunction with UAV-based images to recognize mammals in the African Savannah. This study relates the challenges related to this task and proposes a series of suggestions to overcome them, focusing mostly on imbalances in the labeled dataset. The identification of wild-life, also, was not only performed in terrestrial environments, but also in marine spaces, where a recent publication \citep{Gray2019} implemented a CNN-based semantic segmentation method to identify cetacean species, mainly blue, humpback, and minke whales, in the ocean. These studies not only demonstrate that such methods can be highly accurate at different tasks but also imply the potential of DL approaches for UAVs in the current literature.

\subsection{Urban Mapping}

For urban environments, many DL-based proposals with UAV data have been presented in the literature in the last years. The high-spatial-resolution easily provided by UAV embedded sensors are one of the main reasons behind its usage in these areas. Object detection and instance segmentation methods in those images are necessary to individualize, recognize, and map highly-detailed targets. Thus, many applications rely on CNNs and, in small cases, RNNs (CNN-LSTM) to deal with them. Some of the most common examples encountered in this category during our survey are the identification of pedestrians, car and traffic monitoring, segmentation of individual tree-species in urban forests, detection of cracks in concrete surfaces and pavements, building extraction, etc. Most of these applications were conducted with RGB type of sensors, and, in a few cases, spectral ones. 

The usage of RGB sensors is, as aforementioned, a preferred option for small-budget experiments, but also is related to another important preference of CNNs, and that is that features like pixel-size, form, and texture of an object are essential to its recognition. In this regard, novel experiments could compare the performance of DL-based methods with RGB imagery with other types of sensors. As low-budget systems are easy to implement in larger quantities, many urban monitoring activities could benefit from such investigations. In urban areas, the importance of UAV real-time monitoring is relevant, and that is one of the current objectives when implementing such applications.

The most common practices on UAV-based imagery in urban environments with DL-based methods involve the detection of vehicles and traffic. Car identification is an important task to help urban monitoring and may be useful for real-time analysis of traffic flow in those areas. It is not an easy task, since vehicles can be occluded by different objects like buildings and trees, for example. A recent approach using RGB video footage obtained with UAV, as presented in \citep{Zhang2019}, used an object detection CNN for this task. They also dealt with differences in traffic monitoring to motorcycles, where a frame-by-frame analysis enabled the neural network to determine if the object in the image was a person (pedestrian) or a person riding a motorcycle since differences in its pattern and frame-movement indicated it. Regarding pedestrian traffic, an approach with thermal cameras presented by \citep{DeOliveira2018} demonstrated that CNNs are appropriate to detect persons with different camera rotations, angles, sizes, translation, and scale, corroborating the robustness of its learning and generalization capabilities.

Another important survey in those areas is the detection  and localization of single-tree species, as well as the segmentation of their canopies. Identifying individual species of vegetation in urban locations is an important requisite for urban-environmental planning since it assists in inventorying species and providing information for decision-making models. A recent study \citep{DosSantos2019} applied object detection methods to detect and locate tree-species threatened by extinction. Following their intentions, a research \citep{Torres2020} evaluated semantic segmentation neural networks to map endangered tree-species in urban environments. While one approach aimed to recognize the object to compose an inventory, the other was able to identify it and return important metrics, like its canopy-area for example. Indeed, some proposals that were implemented in a forest type of study could also be adopted in urban areas, and this leaves an open field for future research that intends to evaluate DL-based models in this environment. Urban areas pose different challenges for tree monitoring, so these applications need to consider their characteristics.

DL-based methods have also been used to recognize and extract infrastructure information. An interesting approach demonstrated by \citep{Boonpook2021}, based on semantic segmentation methods, was able to extract buildings in heavily urbanized areas, with unique architectural styles and complex structures. Interestingly enough, a combination of RGB with a DSM improved building identification, indicating that the segmentation model was able to incorporate appropriate information related to the objects' height. This type of combinative approach, between spatial-spectral data and height, may be useful in other identification and recognition approaches. Also regarding infrastructure, another possible application in urban areas is the identification and location of utility poles \citep{Gomes2020}. This application, although being of rather a specific example, is important to maintain and monitor the conditions of poles regularly. These types of monitoring in urban environments is something that benefits from DL-based models approaches, as it tends to substitute multiple human inspection tasks. Another application involves detecting cracks in concrete pavements and surfaces \citep{Bhowmick2020}. Because some regions of civil structures are hard to gain access to UAV-based data with object detection networks may be useful to this task, returning a viable real-life application.

Another topic that is presenting important discoveries relates to land cover pixel segmentation in urban areas, as demonstrated by \citep{Benjdira2019}. In this investigation, an unsupervised domain adaptation method based on GANs was implemented, working with different data from UAV-based systems, while being able to improve image segmentation of buildings, low vegetation, trees, cars, and impervious surfaces. As aforementioned, GANs or DCGANs are quickly gaining the attention of computer vision communities due to their wide area of applications and the way they function by being trained to differentiate between real and fake data \citep{Goodfellow2014}. Regardless, its usage in UAV-based imagery is still underexplored, and future investigations regarding not only land change and land cover but also other types of applications' accuracies may be improved with them. Nonetheless, apart from differences in angles, rotation, scales, and other UAV-based imagery-related characteristics, diversity in urban scenarios is a problem that should be considered by unsupervised approaches. Therefore, in the current state, DL-based networks still may rely on some supervised manner to guide image processing, specifically regarding domain shift factors.

\subsection{Agricultural Mapping}

Precision agriculture applications have been greatly benefited from the integration between UAV-based imagery and DL methods in recent scientific investigations. The majority of issues related to these approaches involve object detection and feature extraction for counting plants and detecting plantation lines, recognizing plantation-gaps, segmentation of plant species and invasive species such as weeds, phenology, and phenotype detection, and many others. These applications offer numerous possibilities for this type of mapping, especially since most of these tasks are still conducted manually by human-vision inspection. As a result, they can help precision farming practices by returning predictions with rapid, unbiased, and accurate results, influencing decision-making for the management of agricultural systems. 

Regardless, although automatic methods do provide important information in this context, they face difficult challenges. Some of these include similarity between the desired plant and invasive plants, hard-to-detect plants in high-density environments (i.e. presenting small spacing between plants and lines), plantation-lines that do not follow a straight-path, edge-segmentation in mapping canopies with conflicts between shadow and illumination, and many others. Still, novel investigations aim to achieve a more generative capability to these networks in dealing with such problems. In this sense, approaches that implement methods in more than one condition or plantation are being the main focus of recent publications. Thus, varied investigation scenarios are currently being proposed, with different types of plantations, sensors, flight-altitudes, angles, spatial and spectral divergences, dates, phenological-stages, etc.

An interesting approach that has the potential to be expanded to different orchards was used in \citep{Apolo-Apolo2020}. There, a low-altitude flight approach was adopted with side-view angles to map yield by counting fruits with the CNN-based method. Counting fruits is not something entirely new in DL-based approaches, some papers demonstrated the effectiveness of bounding-box and point-feature methods to extract it \citep{Biffi2021,Tian2019b,Kang2020b} aside from several differences in occlusion, lightning, fruit size, and image corruption. 

Today's deep networks demonstrate high potential in yield-prediction, as some applications are adapted to CNN architectures mainly because of its benefits in image processing. One of which includes predicting pasture-forage with only RGB images \citep{Castro2020}. Another interesting example in crop-yield estimates is presented by \citep{Nevavuori2020}, where a CNN-LSTM was used to predict yield with a spatial multitemporal approach. There the authors implemented this structure since RNNs are more appropriate to learn from temporal data, while a 3D-CNN was used to process and classify the image. Although used less frequently than CNNs in the literature, there is emerging attention to LSTM architectures in precision agriculture approaches, which appear to be an appropriate intake for temporal monitoring of these areas.

Nonetheless, one of the most used and beneficiated approaches in precision agriculture with DL-based networks is counting and detecting plants and plantation lines. Counting plants is essential to produce estimates regarding production rates, as well as, by geolocating it, determine if a problem occurred during the seedling process by identifying plantation-gaps. In this regard, plantation-lines identification with these gaps is also a desired application. Both object detection and image segmentation methods were implemented in the literature, but most approaches using image semantic segmentation algorithms rely on additional procedures, like using a blob detection method \citep{Kitano2019}, for example. These additional steps may not always be desirable, and to prove the generality capability of one model, multiple tests at different conditions should be performed. 

For plantation-line detection, segmentations are currently being implemented and often used to assist in more than one information extraction. In \citep{Osco2021a} semantic segmentation methods were applied in UAV-based multispectral data to extract canopy areas and was able to demonstrate which spectral regions were more appropriate to it. A recent application with UAV-based data was also proposed in \citep{Osco2021b}, where a CNN model is presented to simultaneously count and detect plants and plantation-lines. This model is based on a confidence map extraction and was an upgraded version from previous research with citrus-tree counting \citep{Osco2020}. This CNN works by implementing some convolutional layers, a Pyramid Pooling Module (PPM) \citep{Zhao2017}, and a Multi-Stage Module (MSM) with two information branches that, concatenated at the end of the MSM processes, shares knowledge learned from one to another. This method ensured that the network learned to detect plants that are located at a plantation-line, and understood that a plantation-line is formed by linear conjunction of plants. This type of method has also been proved successful in dealing with highly-dense plantations. Another research \citep{Ampatzidis2019} that aimed to count citrus-trees with a bounding-box-based method also returned similar accuracies. However, it was conducted in a sparse plantation, which did not impose the same challenges faced at \citep{Osco2020,Osco2021b}. Regardless, to deal with highly dense scenes, feature extraction from confidence maps appears to be an appropriate approach.

However, agricultural applications do not always involve plant counting or plantation-line detection. Similar to wild-animal identification as included in other published studies \citep{Kellenberger2020,Gray2019}, there is also an  interest in cattle detection, which is still an onerous task for human-inspection. In UAV-based imagery, some approaches included  DL-based bounding-boxes methods \citep{Barbedo2019}, which were also successfully implemented. DNNs used for this task are still underexplored, but published investigations \citep{Rivas2018} argue that  one of the main reasons behind the necessity to use DL methods is based on occurrences of changes in terrain (throughout the seasons of the year) and the non-uniform distribution of the animals throughout the area. On this matter, one interesting approach should involve the usage of real-time object detection on the flight. This is because it is difficult to track animal movement, even in open areas such as pastures, when a UAV system is acquiring data. Another agricultural application example refers to the monitoring offshore aquaculture farms using UAV-underwater color imagery and DL models to classify them \citep{Bell2020}. These examples reveal the widespread variety of agriculture problems that can be attended with the integration of DL models and UAV remote sensing data.

Lastly, a field yet to be also explored in the literature is the identification and recognition of pests and disease indicators in plants using DL-based methods. Most recent approaches aimed to identify invasive species, commonly named ``weeds'', in plantation-fields. In a demonstration with unsupervised data labeling, \citep{DianBah2018} evaluated the performance of a CNN-based method to predict weeds in the plantation lines of different crops. This pre-processing step to automatically generate labeled data, which is implemented outside the CNN model structure, is an interesting approach. However, others prefer to include a ``one-step'' network to deal with this situation, and different fronts are emerging in the literature. Unsupervised domain adaptation, in which the network extracts learning features from new unviewed data, is one of the most current aimed models. 

A recent publication \citep{LI2020105745} proposed it to recognize and count in-field cotton-boll status identification. Regardless, with UAV-based data examples, this is still an issue. As for disease detection, a study \citep{Kerkech2020} investigated the use of image segmentation for vine-crops with multispectral images, and was able to separate visible symptoms (RGB), infrared symptoms (i.e. when considering only the infrared band) and in an intersection between visible and infrared spectral data. Another interesting example regarding pests identification with UAV-based image was demonstrated in \citep{Tetila2020} where superpixel image samples of multiple pest species were considered, and activation filters used to recognize undesirable visual patterns implemented alongside different DL-based architectures.

\section{Publicly Available UAV-Based Datasets}

As mentioned, one of the most important characteristics of DL-based methods is that they tend to increase their learning capabilities as a number of labeled examples are used to train a network. In most of the early approaches to remote sensing data, CNNs were initialized with pre-trained weights from publicly available image repositories over the internet. However, most of these repositories are not from data acquired with remote sensing platforms. Still, there are some known aerial repositories with labeled examples, which were presented in recent years, such as the DOTA \citep{Xia2018}, UAVDT \citep{Du2018}, VisDrone \citep{B2019}, WHU-RS19 \citep{Sheng2012}, RSSCN7 \citep{Zou2015b}, RSC11 \citep{Zhao2016}, Brazilian Coffee Scene \citep{Penatti2015} datasets. These and others are gaining notoriety in UAV-based applications and could be potentially used to pre-train or benchmark DL methods. These datasets not only serve as an additional option to start a network but also may help in novel proposals to be compared against the evaluated methods.

Since there is a still scarce amount of labeled examples with UAV-acquired data, specifically in multispectral and hyperspectral data, we aimed to provide UAV-based datasets in both urban and rural scenarios for future research to implement and compare the performance of novel DL-based methods with them. Table~\ref{tab} summarizes some of the information related to these datasets, as well as indicates recent publications in which previously conducted approaches were implemented, as well as the results achieved on them. They are available on the following webpage, which is to be constantly updated with novel labeled datasets from here on: \href{https://sites.google.com/view/geomatics-and-computer-vision/home/datasets}{Geomatics and Computer Vision/Datasets}

\begin{table*}[ht!]
\centering
\caption{\small \centering UAV-based datasets that are publically available from previous research.\label{tab}}
\resizebox{14cm}{!}{%
\begin{tabular}{l l l l l l l}
\hline
\textbf{Reference}	& \textbf{Task}	& \textbf{Target} & \textbf{Sensor} & \textbf{GSD\textsubscript{(cm)}} & \textbf{Best Method} & \textbf{Result}\\
\hline
\citep{DosSantos2019}	& Detection & Trees	& RGB & 0.82 & RetinaNet & AP = 92.64\%\\
\citep{Torres2020} & Segmentation & Trees & RGB & 0.82 & FC-DenseNet & F1 = 96.0\%\\
\citep{Osco2021a} & Segmentation & Citrus & Multispectral & 12.59 & DDCN & F1 = 94.4\%\\
\citep{Osco2021b} & Detection & Citrus & RGB & 2.28 & \citep{Osco2021b} & F1 = 96.5\%\\
\citep{Osco2021b} & Detection & Corn & RGB & 1.55 & \citep{Osco2021b} & F1 = 87.6\%\\
\citep{Osco2020} & Detection & Citrus & Multispectral & 12.59 & \citep{Osco2020} & F1 = 95.0\%\\
\hline
\end{tabular}}
\end{table*}

\section{Perspectives in Deep Learning with UAV Data}

There is no denying that DL-based methods are a powerful and important tool to deal with the numerous amounts of data daily produced by remote sensing systems. What follows in this section is a short commentary on the near perspectives of one of the most emerging fields in the DL and remote sensing communities that could be implemented with UAV-based imagery. These topics, although individually presented here, have the potential to be combined, as already performed in some studies, contributing to the development of novel approaches.

In general, DL architectures require low resolution input images (e.g., $512 \times 512$ pixels). High resolution images are generally scaled to the size required for processing. However, UAVs have the advantage of capturing images in higher resolution than most other types of sensing platforms aside from proximal sensing, and the direct application of traditional architectures may not take advantage of this feature. As such, processing images with DL while maintaining high resolution in deeper layers is a challenge to be explored. In real-time applications, such as autonomous navigation, this processing must be fast, which opens up a range of research related to reducing the complexity of architectures while preserving accuracy. Regarding DL, recently, some CNN architectures that try to maintain high resolution in deeper layers, such as HRNet, have been proposed \citep{Kannojia2018}. These novel architectures can really take advantage of the high resolution from UAV images compared to commonly available orbital data.

To summarize, the topics addressed in this section compose some of the hot topics in the computer vision community, and the combination of them with remote sensing data can contribute to the development of novel approaches in the context of UAV mapping. In this regard, it is important to emphasize that not only these topics are currently being investigated by computer vision research, but that they also are being fastly implemented in multiple approaches aside from remote sensing. As other domains are investigated, novel ways of improving and adapting these networks can be achieved. Future studies in remote sensing communities, specifically on UAV-based systems, may benefit from these improvements and incorporate them into their applications.

\subsection{Real-Time Processing}

Most of the environmental, urban, and agricultural applications presented in this study can benefit from real-time responses. Although UAV and DL-based combinations speed up the processing pipeline, these algorithms are highly computer-intensive. Usually, they do require post-processing in data centers or dedicated Graphics Processing Units (GPUs) machines. Although DL is considered a fast method to extract information from data after its training, it still bottlenecks real-time applications mainly because of the number of layers intrinsic to the DL methods architecture. Research groups, especially from the IoT industry/academy, race to develop real-time DL methods because of it. The approach usually goes in two directions: developing faster algorithms and developing dedicated GPU processors.

DL models use 32-bit floating points to represent the weights of the neural network. A simple strategy known as quantization reduces the amount of memory required by DL models representing the weights, using 16, 8, or even 1 bit instead of 32-bits floating points. A 32-bit full precision ResNet-18 \citep{He2016} achieves 89.2\% top-5 accuracy on the ImageNet dataset \citep{imagenet2018imagenet}, while the ResNet-18 \citep{He2016} ported to XNOR-Net achieves 73.2\% top-5 accuracy in the same dataset. The quantization goes beyond weights, in all network components, while the literature reports activation functions and gradient optimizations quantized methods. The survey conducted in \citep{guo2018survey} gives an important overview of quantization methods. Also, knowledge distillation \citep{hinton2015distilling} is another example of a training model using a smaller network, where a larger ``teacher'' network guides the learning process of a smaller ``student'' network.

Another strategy to develop fast DL models is to design layers with fewer parameters that are still capable of retaining predictive performance. MobileNets \citep{howard2017mobilenets} and its variants are a good example of this idea. In specific tasks, such as object detection, it is possible to develop architectural enhancements for this approach, such as the Context Enhanced Module (CEM) and the Spatial Attention Module (SAM) \citep{qin2019thundernet}. When considering even smaller computational power, it is possible to find DL running on microcontroller units (MCU) where the memory and computational power are 3-4 orders of magnitude smaller than mobile phones.

On hardware, the industry has already developed embedded AI platforms that run DL algorithms. NVIDIA's Jetson is amongst the most popular choices and a survey \citep{mittal2019survey} of studies using the Jetson platform and its applications demonstrate it. Also, a broader survey on this theme, that considers GPU, ASIC, FPGA, and MCUs of AI platforms, can be read in \citep{imran2020embedded}. Regardless, research in the context of UAV remote sensing is quite limited, and there is a gap that can be fulfilled by future works. Several applications can be benefited by this technology, including, for example, agricultural spraying UAV, which can recognize different types of weeds in real-time, and simultaneously use the spray. Other approaches may include real-time monitoring of trees in both urban and forest environments, as well as the detection of other types of objects that benefit from a rapid intake.

\subsection{Dimensionality Reduction}

Due to recent advances in capture devices, hyperspectral images can be acquired even in UAVs.
 These images consist of tens to hundreds of spectral bands that can assist in the classification of objects in a given application. However, two main issues arise from the high dimensionality: i) the bands can be highly correlated, and ii) the excessive increase in the computational cost of DL models. High-dimensionality could invoke a problem known as the Hughes phenomenon, which is also known as the curse of dimensionality, i.e., when the accuracy of a classification is reduced due to the introduction of noise and other implications encountered in hyperspectral or high-dimensional data  \citep{Hennessy2020}. Regardless, hyperspectral data may pose an hindrance for the DL-based approaches accuracies, thus being an important issue to be considered in remote sensing practices. The classic approach to address high dimensionality is by applying a Principal Component Analysis (PCA) \citep{Licciardi2012}.
 
Despite several proposals, PCA is generally not applied in conjunction with DL, but as a pre-processing step. Although this method may be one of the most known approaches to reduce dimensionality when dealing with hyperspectral data, different intakes were already presented in the literature. A novel DL approach, implemented with UAV-based imagery, was demonstrated by Miyoshi et al. \citep{Miyoshi2020}. There, the authors proposed a one-step approach, conducted within the networks' architecture, to consider a combination of bands of a hyperspectral sensor that were highly related to the labeled example provided in the input layer at the initial stage of the network. Another investigation \citep{VADDI2020103457} combines a band selection approach, spatial filtering, and CNN to simultaneously extract the spectral and spatial features. Still, the future perspective to solve this issue appears to be a combination of spectral band selection and DL methods in an end-to-end approach. Thus, both selection and DL methods can exchange information and improve results. This can also contribute to understanding how DL operates with these images, which was slightly accomplished at Miyoshi et al. \citep{Miyoshi2020}.  

\subsection{Domain Adaptation and Transfer Learning}

The training steps of DL models are generally carried out on images captured in a specific geographical region, in a short-time period, or on single capture equipment (also known as domains). When the model is used in practice, it is common for spectral shifts to occur between the training and test images due to differences in acquisition, geographic region, atmospheric conditions, among others \citep{TuiaGRSM2016}. Domain adaptation is a technique for adapting models trained in a source domain to a different, but still related, target domain. Therefore, domain adaptation is also viewed as a particular form of transfer learning \citep{TuiaGRSM2016}. On the other hand, transfer learning \citep{zhuang2020comprehensive,tan2018survey} does include applications in which the characteristics of the domain's target space may differ from the source domain.

A promising research line for domain adaptation and transfer learning is to consider GANs \citep{Goodfellow2014,ElshamliJSTAEORS2017}. For example, \citep{BenjdiraRS2019} proposed the use of GANs to convert an image from the source domain to the target domain, causing the source images to mimic the characteristics of the images from the target domain. Recent approaches seek to align the distribution of the source and target domains, although they do not consider direct alignment at the level of the problem classes. Approaches that are attentive to class-level shifts may be more accurate, as the category-sensitive domain adaptation proposed by \citep{FangRS2019}. Thus, these approaches reduce the domain shift related to the quality and characteristics of the training images and can be useful in practice for UAV remote sensing.

\subsection{Attention-Based Mechanisms}

Attention mechanisms aim to highlight the most valuable features or image regions based on assigning different weights for them in a specific task. It is a topic that has been recently applied in remote sensing, providing significant improvements. As pointed out by \citep{XuRS2018}, high-resolution images in remote sensing provide a large amount of information and exhibit minor intra-class variation while it tends to increase. These variations and a large amount of information make extraction of relevant features more difficult, since traditional CNNs process all regions with the same weight (relevance). Attention mechanisms, such as the one proposed by \citep{XuRS2018}, are useful tools to focus the feature extraction in discriminative regions of the problem, be it image segmentation \citep{DingTGRS2021,SuIGARSS2019,ZhouSensors2020}, scene-wise classification \citep{ZhuRS2019,LiRS2020}, or object detection \citep{LiICIP2019,LiRS2020}, as others.

Besides, \citep{SuIGARSS2019} argue that when remote sensing images are used, they are generally divided into patches for training the CNNs. Thus, objects can be divided into two or more sub-images, causing the discriminative and structural information to be lost. Attention mechanisms can be used to aggregate learning by focusing on relevant regions that describe the objects of interest, as presented in \citep{SuIGARSS2019}, through a global attention upsample module that provides global context and combines low and high-level information. Recent advances in computer vision were achieved with attention mechanisms for classification (e.g., Vision Transformer \citep{dosovitskiy2020image} and Data-efficient Image Transformers \citep{touvron2020training}) and in object detection (e.g., DETR \citep{CarionECCV2020}) that have not yet been fully evaluated in remote sensing applications. Some directions also point to the use of attention mechanisms directly in a sequence of image patches \citep{dosovitskiy2020image,touvron2020training}. These new proposals can improve the results already achieved in remote sensing data, just as they have advanced the results on the traditional image datasets in computer vision (e.g., ImageNet \citep{imagenet2018imagenet}).

\subsection{Few-Shot Learning}

Although recent materials demonstrated the feasibility of DL-based methods for multiple tasks, they still are considered limited in terms of high generalization. This occurs when dealing with the same objects in different geographical areas or when new object classes are considered. Traditional solutions require retraining the model with a robust labeled dataset for the new area or object. Few-shot learning aims to cope with situations in which few labeled datasets are available. A recent study \citep{Li2020TGRS}, in the context of scene classification, pointed out that few-shot methods in remote sensing are based on transfer learning and meta-learning. Meta-learning can be more flexible than transfer learning, and when applied in the training set to extract meta-knowledge, contributes significantly to few-shot learning in the test set. An interesting strategy to cope with large intraclass variation and interclass similarity is the implementation of the attention mechanism in the feature learning step, as previously described. The datasets used in the \citep{Li2020TGRS} study were not UAV-based; however, the strategy can be explored in UAV imagery.

In the context of UAV remote sensing, there are few studies on few-shot learning. Recently, an investigation \citep{Karami2020} aimed for the detection of maize plants using the object detection method CenterNet. The authors adopted a transfer learning strategy using pre-trained models from other geographical areas and dates. Fewer images (in total, 150 images), when compared to the previous training (with 600 images), from the new area were used for fine-tuning the model. Based on the literature survey, there is a research-gap to be further explored in the context of object detection using few-shot learning in UAV remote sensing. The main idea behind this is to consider less labeled datasets for training, which may help in some remote applications where data availability is scarce or presents few occurrences. 

\subsection{Semi-Supervised Learning and Unsupervised Learning}

With the increasing availability of remote sensing images, the labeling task for supervised training of DL models is expensive and time-consuming. Thus, the performance of DL models is impacted due to the lack of large amount of labeled training images. Efforts have been made to consider unlabeled images in training through unsupervised (unlabeled images only) and semi-supervised (labeled and unlabeled images) learning. In remote sensing, most semi-supervised or unsupervised approaches are based on transfer learning, which usually requires a supervised pre-trained model \citep{Liu2020}. In this regard, a recent study \citep{Kang2020} proposed a promising approach for unlabeled remote sensing images that define spatial augmentation criteria for relating close sub-images. Regardless, this is still an underdeveloped practice with UAV-based data and should be investigated in novel approaches.

Future perspectives point to the use of contrastive loss \citep{BachmanNIPS2019,Tian2019,hjelm2019learning,HeCVPR2020} and clustering-based approaches \citep{CaronECCV2018,caron2021unsupervised}. Recent publications have shown interesting results with the use of contrastive loss that has not yet been fully evaluated in remote sensing. For example, \citep{HeCVPR2020} proposed an approach based on contrastive loss that surpassed the performance of its supervised pre-trained counterpart. As for clustering-based methods, they often group images with similar characteristics \citep{CaronECCV2018}. On this matter, a research \citep{CaronECCV2018} presented an approach that groups the data while reinforcing the consistency between the cluster assignments produced for a pair of images (same images with two augmentations). An efficient and effective way to use a large number of unlabeled images can considerably improve the performance, mainly related to the generalizability of the models.

\subsection{Multitask Learning}

Multitask learning aims to perform multiple tasks simultaneously. Several advantages are mentioned in \citep{crawshaw2020}, including fast learning and the minimization of overfitting problems. Recently, in the context of UAV remote sensing, there were some important researches already developed. A study \citep{WANG2021} proposed a method to conduct three tasks (semantic segmentation, height estimation, and boundary detection), which also considered boundary attention modules. Another research \citep{Osco2021b} simultaneously detecting plants and plantation lines in UAV-based imagery. The proposed network benefited from the contributions of considering both tasks in the same structure, since the plants must, essentially belong to a plantation line. In short, improvements occurred in the detection task when line detection was considered at the same time. This approach can be further explored in several UAV-based remote sensing applications.

\subsection{Open-Set}

The main idea of an open-set is to deal with unknown or unseen classes during the inference in the testing set \citep{Bendale_2016_CVPR}. As the authors mention, recognition in real-world scenarios is ``open-set'', different from neural networks’ nature, which is in a ``close-set''. Consequently, the testing set is classified considering only the classes used during the training. Therefore, unknown or unseen classes are not rejected during the test. There are few studies regarding open-set in the context of remote sensing. Regarding semantic segmentation of aerial imagery, a study by \citep{Silva2020} presented an approach considering the open-set context. There, an adaptation of a close-set semantic segmentation method, adding a probability threshold after the softmax, was conducted. Later, a post-processing step based on morphological filters was applied to the pixels classified as unknown to verify if they are inside pixels or from borders. Another interesting approach is to combine open-set and domain adaptation methods, as proposed by \citep{Adayel_2020} in the remote sensing context.

\subsection{Photogrammetric Processing}

Although not as developed as other practices, DL-based methods can be adopted for processing and optimizing the UAV photogrammetric processing task. This process aims to generate a dense point cloud and an orthomosaic, and it is based on Structure-from-Motion (SfM) and Multi-View Stereo (MVS) techniques. In SfM, the interior and exterior orientation parameters are estimated, and a sparse point cloud is generated. A matching technique between the images is applied in SfM. A recent survey on image matching \citep{Ma2021} concluded that this thematic is still an open problem and pointed out the potential of DL is this task. The authors mentioned that DL techniques are mainly applied to feature detection and description, and further investigations on feature matching can be explored. Finally, they pointed out that a promising direction is the customization of modern feature matching techniques to attend SfM. 

Regarding DL for UAV image matching, there is a lack of work indicating a potential for future exploration.  In the UAV photogrammetric process, DL also can be used in filtering the DSM, which is essential to generate high-quality orthoimages. Previous work \citep{GEVAERT2018} showed the potential of using DL to filter the DSM and generate the DTM. Further investigations are required in this thematic, mainly considering UAV data. Besides, another task that can be beneficiated by DL is the color balancing between images when generating orthomosaic from thousands of images, corresponding to extensive areas.

\section{Conclusions}

DL is still considered up to the time of writing, a ``black-box'' type of solution for most of the problems, although novel research is advancing in minimizing this notion at considerable proportions. Regardless, in the remote sensing domain, it already provided important discoveries on most of its implementations. Our literature revision has focused on the application of these methods in UAV-based image processing. In this sense, we structured our study to offer more of a comprehensive approach to the subject while presenting an overview of state-of-the-art techniques and perspectives regarding its usage. As such, we hope that this literature revision may serve as an inclusive survey to summarize the UAV applications based on DNNs. Thus, in the evaluated context, this review concludes that:
\begin{enumerate}
\item In the context of UAV remote sensing,  most of the published materials are based on object detection methods and RGB sensors; however, some applications, as in precision agriculture and forest-related, benefit from multi/hyperspectral data;
\item There is a need for additional labeled public available datasets obtained with UAVs to be used to train and benchmark the networks. In this context, we contributed by providing a repository with some of our UAV datasets in both agricultural and environmental applications;
\item Even though CNNs are the most adopted architecture, other methods based on CNN-LSTMs and GANs are gaining attention in UAV remote sensing and image applications, and future UAV remote sensing works may benefit from their inclusion;
\item DL, when assisted by GPU processing, can provide fast inference solutions. However there is still a need for further investigation regarding real-time processing using embedded systems on UAVs, and, lastly;
\item Some promising thematics, such as open-set, attention-based mechanisms, few shot and multitask learning can be combined and provide novel approaches in the context of UAV remote sensing; also, these thematics can contribute significantly to the generalization capacity of the DNNs.
\end{enumerate}


\section*{Acknowledgements}

This study was financed in part by the Coordenação de Aperfeiçoamento de Pessoal de Nível Superior (CAPES) - Finance Code 001. The authors are funded by the Support Foundation for the Development of Education, Science, and Technology of the State of Mato Grosso do Sul (FUNDECT; 71/009.436/2022) and the Brazilian National Council for Scientific and Technological Development (CNPq; 433783/2018-4, 310517/2020-6; 405997/2021-3; 308481/2022-4; 305296/2022-1).

\section*{Conflicts of Interest}

The authors declare no conflict of interest. The funders had no role in the design of the study; in the collection, analyses,or interpretation of data; in the writing of the manuscript, or in the decision to publish the results.

\pagebreak
\section*{Abbreviations}{
The following abbreviations are used in this manuscript:\\

\noindent 
\begin{tabular}{@{}ll}
AdaGrad & Adaptive Gradient Algorithm\\
AI & Artificial Intelligence\\
ANN & Artificial Neural Network\\
CEM & Context Enhanced Module\\
CNN & Convolutional Neural Network\\
DCGAN & Deep Convolutional Generative Adversarial network\\
DDCN & Deep Dual-domain Convolutional neural Network\\
DL & Deep Learning\\
DNN & Deep Neural Network\\
DEM & Digital Elevation Model\\
DSM & Digital Surface Model\\
FPS & Frames per Second\\
GAN & Generative Adversarial Network\\
GPU & Graphics Processing Unit\\
KL & Kullback-Leibler\\
LSTM & Long Short-Term Memory\\
IoU & Intersection over Union\\
ML & Machine Learning\\
MAE & Mean Absolute Error\\
MAPE & Mean Absolute Percentage Error\\
MRE & Mean Relative Error\\
MSE & Mean Squared Error\\
MSLE & Mean Squared Logarithmic Error\\
MSM & Multi-Stage Module\\
MVS & Multiview Stereo\\
NAS & Network Architecture Search\\
PCA & Principal Component Analysis\\
PPM & Pyramid Pooling Module\\
r & Correlation Coefficient\\
RMSE & Root Mean Squared Error\\
RNN & Recurrent Neural Network\\
ROC & Receiver Operating Characteristics\\
RPA & Remotely Piloted Aircraft\\
SAM & Spatial Attention Module\\
SGD & Stochastic Gradient Descent\\
SfM & Structure from Motion\\
UAV & Unmanned Aerial Vehicle\\
WOS & Web of Science
\end{tabular}
}

\normalsize


\end{document}